\pdfoutput=1

\documentclass[11pt]{article}

\usepackage{acl}

\usepackage{times}
\usepackage{latexsym}

\usepackage[T1]{fontenc}

\usepackage[utf8]{inputenc}

\usepackage{microtype}

\usepackage{graphicx}
\usepackage{amsmath}
\usepackage{algorithm}
\usepackage{algpseudocode}
\usepackage{enumitem}
\usepackage{subfig}

\newcommand{\mxnlg}{Meta-X\textsubscript{NLG}}
\usepackage{xurl}

\title{Meta-X\textsubscript{NLG}: A Meta-Learning Approach Based on Language Clustering for Zero-Shot Cross-Lingual Transfer and Generation}

\author{Kaushal Kumar Maurya \\
  Indian Institute of Technology Hyderabad \\
  Hyderabad, India \\
  \texttt{cs18resch11003@iith.ac.in} \\\And 
  \hspace{10mm} Maunendra Sankar Desarkar \\
  \hspace{10mm} Indian Institute of Technology Hyderabad \\
  \hspace{10mm} Hyderabad, India \\
  \hspace{10mm} \texttt{maunendra@cse.iith.ac.in} \\}

\begin{document}
\maketitle
\begin{abstract}
Recently, the NLP community has witnessed a rapid advancement in multilingual and cross-lingual transfer research where the supervision is transferred from high-resource languages (HRLs) to low-resource languages (LRLs). However, the cross-lingual transfer is not uniform across languages, particularly in the zero-shot setting. Towards this goal, one promising research direction is to learn shareable structures across multiple tasks with limited annotated data. The downstream multilingual applications may benefit from such a learning setup as most of the languages across the globe are low-resource and share some structures with other languages. In this paper, we propose a novel meta-learning framework (called \mxnlg) to learn shareable structures from \textit{typologically} diverse languages based on \textit{meta-learning} and \textit{language clustering}. This is a step towards uniform cross-lingual transfer for unseen languages. We first cluster the languages based on language representations and identify the centroid language of each cluster. Then, a meta-learning algorithm is trained with all centroid languages and evaluated on the other languages in the zero-shot setting. We demonstrate the effectiveness of this modeling on two NLG tasks (Abstractive Text Summarization and Question Generation), 5 popular datasets and 30 typologically diverse languages. Consistent improvements over strong baselines demonstrate the efficacy of the proposed framework. The careful design of the model makes this end-to-end NLG setup less vulnerable to the accidental translation problem, which is a prominent concern in zero-shot cross-lingual NLG tasks.
\end{abstract}

\section{Introduction}
\label{sec:intro}
There are more than 7000 known living languages across the globe. 95\% of the world’s population does not speak English as their first language and 75\% does not speak English at all\footnote{\url{https://www.ethnologue.com/statistics}}. Most of the languages are low-resource languages as they do not have adequate resources for natural language processing research \cite{joshi-etal-2020-state}. On the other hand, a vast majority of studies in NLP research are conducted on English data \cite{bender2019benderrule}. To democratize the NLP research for the benefit of the large global community, it is essential to focus on the non-English languages. However, creating/collecting task-specific annotated data for all the languages is expensive and time-consuming. Moreover, human languages are dynamic as new words and domains are added continuously. An alternate solution is to investigate NLP modeling techniques that allow to train the model with high-resource languages like English and transfer supervision to low-resource languages (with limited annotated data) or unseen languages for several NLP applications. Recently, there has been promising progress on cross-lingual transfer learning research \cite{hu2020xtreme, artetxe-etal-2020-cross} but supervision transfer is uneven across languages which leads to large performance gaps. Such performance gaps are observed because models do not account for cultural and linguistic differences in the modeling \cite{DBLP:journals/corr/abs-1909-07009, blasi2021systematic}. This paper is a step towards bridging this gap via meta-learning and language clustering. 

Meta-learning or \textit{learning to learn} \cite{bengio1990learning} is a learning paradigm where the model is trained on diverse tasks and quickly adapts to new tasks given a handful of examples. It has emerged as a promising technique for Machine Learning \cite{finn2017model, koch2015siamese}, Natural Language Understanding \cite{murty2021dreca, yan2020multi} and Machine Translation \cite{gu-etal-2018-meta} tasks.  This work - to the best of our knowledge - is the first attempt to study  \textit{meta-learning techniques for cross-lingual natural language generation (X\textsubscript{NLG})}. Particularly, we focus on zero-shot X\textsubscript{NLG} for low-resource languages. Unlike NLU tasks, we observe that zero-shot NLG is a more challenging setup as text should be generated in unseen languages (which often suffers from accidental translation (AT) problem \cite{xue-etal-2021-mt5}) and is expected to be grammatically coherent, semantically correct and fluent. We aim to address the following research problem: \textit{Does meta-learning algorithm trained on typologically diverse languages (as training task) provide language-agnostic initialization for the zero-shot cross-lingual generation?} 
Our main contributions in this work are listed below:
\begin{itemize}[leftmargin=*]
    \itemsep-0.3em 
    \item We propose \mxnlg\footnote{code \& pre-trained models: \url{https://github.com/kaushal0494/Meta_XNLG}}, a framework for effective cross-lingual transfer and generation based on Model-Agnostic Meta-Learning (MAML) algorithm.
    \item We use language clustering to identify a set of meta-training languages, which provides a more uniform cross-lingual transfer to unseen languages.
    \item We test \mxnlg~ on two NLG tasks (Abstractive Text Summarization and Question Generation), five popular datasets (XL-Sum, Wikilingua, MLQA, TyDiQA and XQuAD) and 30 languages. We observe consistent improvement over strong baselines involving mT5.
    \item We show an effective zero-shot X\textsubscript{NLG} modeling setup, which is less vulnerable to the accidental translation problem.
\end{itemize}

\section{Related Work}
\label{sec:rel_work}
We focus on two threads of related work in this section: (1) cross-lingual generation and  (2) meta-learning for NLP. Traditional approaches for cross-lingual generation use machine translation (MT) in the modelling pipeline \cite{wan2010cross, 8370729, duan-etal-2019-zero}. Such approaches have an inherent problem as translations are generally error-prone. The errors are more when at least one of the languages involved in the translation is a low-resource language. Recently cross-lingual transfer approaches are gaining attention. These methods use parallel data \cite{chi2020cross} and small annotated datasets \cite{kumar-etal-2019-cross} in zero-shot and few-shot cross-lingual generation respectively. \citet{lewis2020pretraining} fine-tune a pre-trained model with multiple low-resource languages and evaluate on a single target language in zero-shot setting. In the same line of research, \citet{DBLP:conf/acl/MauryaDKD21} modified mBART pre-trained model with an unsupervised dataset involving monolingual data in three languages for cross-lingual transfer. This model, called ZmBART, is tested on a small set of languages - English, Hindi and Japanese. Moreover, it has been observed that such cross-lingual transfers are not uniform across the languages \cite{lin-etal-2019-choosing,  blasi2021systematic}. We make an attempt to bridge this gap via meta-learning.

Recently, meta-learning has been actively applied for many NLP applications \cite{bansal2020self, gao-etal-2019-fewrel} and also for NLU tasks such as text classification \cite{van-der-heijden-etal-2021-multilingual}, NER \cite{wu2020enhanced}, task-oriented dialogue and QA \cite{mhamdi-etal-2021-x}, etc. \citet{DBLP:conf/eacl/TaruneshKKRJ21} propose joint meta-learning approach on multiple languages and tasks from XTREME benchmark \cite{hu2020xtreme}. Close to our work, \citet{DBLP:conf/emnlp/NooralahzadehBB20} propose a meta-learning approach for cross-lingual transfer on NLI and QA, both NLU tasks. The authors use one or two randomly selected languages for meta-training. In contrast, we provide a systematic approach based on language clustering to identify the right meta-training languages. Moreover, to the best of our knowledge, ours is the first effort that employs meta-learning for natural language generation. 

\section{Meta-Learning Algorithm: MAML}
\label{sec:maml}
Meta-learning tries to learn structure among multiple tasks such that the new tasks are adapted quickly given few training instances. Among several meta-learning algorithms, we focus on optimization-based algorithms, i.e., Model Agnostic Meta-Learning (MAML) \cite{finn2017model} due to its recent success in multiple NLP and computer vision tasks. MAML progresses in two phases: \textit{meta-training} and \textit{adaptation}. In the meta-training phase, the model learns a good initialization of parameter values by repeatedly simulating the learning process on training tasks. In the adaptation phase, these learned parameters are quickly adapted to new tasks. The underlying constraint is that \textit{all tasks should share some common structure (or come from a task distribution)}. The world's different languages follow this constraint as they came into existence with a common goal of communication, and share some structure. 
For meta-learning purposes, we treat them as different tasks.

Unlike traditional machine learning, meta-learning has \textit{meta-train} and \textit{meta-test} data splits for meta-training and adaptation respectively. Each split consists of tasks that are sampled from  a distribution $p\mathcal{(D)}$ over task datasets $\{\mathcal{D}_1, \mathcal{D}_2, \dots, \mathcal{D}_n\}$ where $\mathcal{D}_i$ is associated with $i^{th}$ task $\mathcal{T}_i$. Each $\mathcal{D}_i$ has \textit{support set} and \textit{query set} $\mathcal{D}_i = \{\mathcal{S}_i, \mathcal{Q}_i \}$. \textit{Support set} and \textit{Query set} are analogous to train and test splits of the traditional machine learning. We use $f_{\theta}$ to denote a neural network model parameterized by $\theta$. 

Meta-training has two-levels of optimization: \textit{inner-loop optimization} and \textit{outer-loop optimization}. In the inner-loop optimization, for each sampled task $\mathcal{T}_i$, the task-specific model parameters $\theta_i^{m}$ are updated by $m$ iterations of stochastic gradient decent (SGD) with support set $\mathcal{S}_i$. The overall model parameters $\theta$ are learned to optimize the performance of models $f_{\theta_i^{(m)}}$ on query sets $\mathcal{Q}_i$ across datasets $p\mathcal{(D)}$ in the outer-loop optimization.
The MAML \cite{finn2017model} objective is:

\begin{equation}
    \theta^* = \arg\min_{\theta} \sum_{D_i \sim p(D)} \mathcal{L}_i(f_{\theta_i^{(m)}})
\end{equation}

 where $\mathcal{L}_i(f_{\theta_i^{(m)}})$ is the loss obtained on query set for task  $\mathcal{T}_i$  and $f_{\theta_i^{(m)}}$ is  obtained after $m$ iteration of SGD update with Task $\mathcal{T}_i$ as: 
 \begin{equation*}
 f_{\theta_i^{(m)}} =  f_{\theta} - \alpha \nabla_{\theta}\mathcal{L}_i(f_{\theta})
 \end{equation*}
In outer-loop optimization, MAML performs \textit{MetaUpdate} which a batch as:
\begin{equation}
    \theta = \theta - \beta \nabla_{\theta} \sum_{D_i \sim p(D)}  \mathcal{L}_i(f_{\theta_i^{(m)}})
\end{equation}

Where $\alpha$ is inner-loop learning rate and $\beta$ is meta (outer-loop) learning rate. 
In the adaptation phase, the model is initialized with with learned optimal  meta-parameters ${\theta}^*$, which is updated by a few steps of SGD with a support set (aka. few-shot learning) and directly evaluated on the query set of the meta-test dataset. Our aim is to perform zero-shot evaluation, so we skip the adaptation phase and directly evaluate the learned model on meta-test datasets.  
\section{Methodology}
\begin{figure*}
    \centering
    \includegraphics[width=12cm]{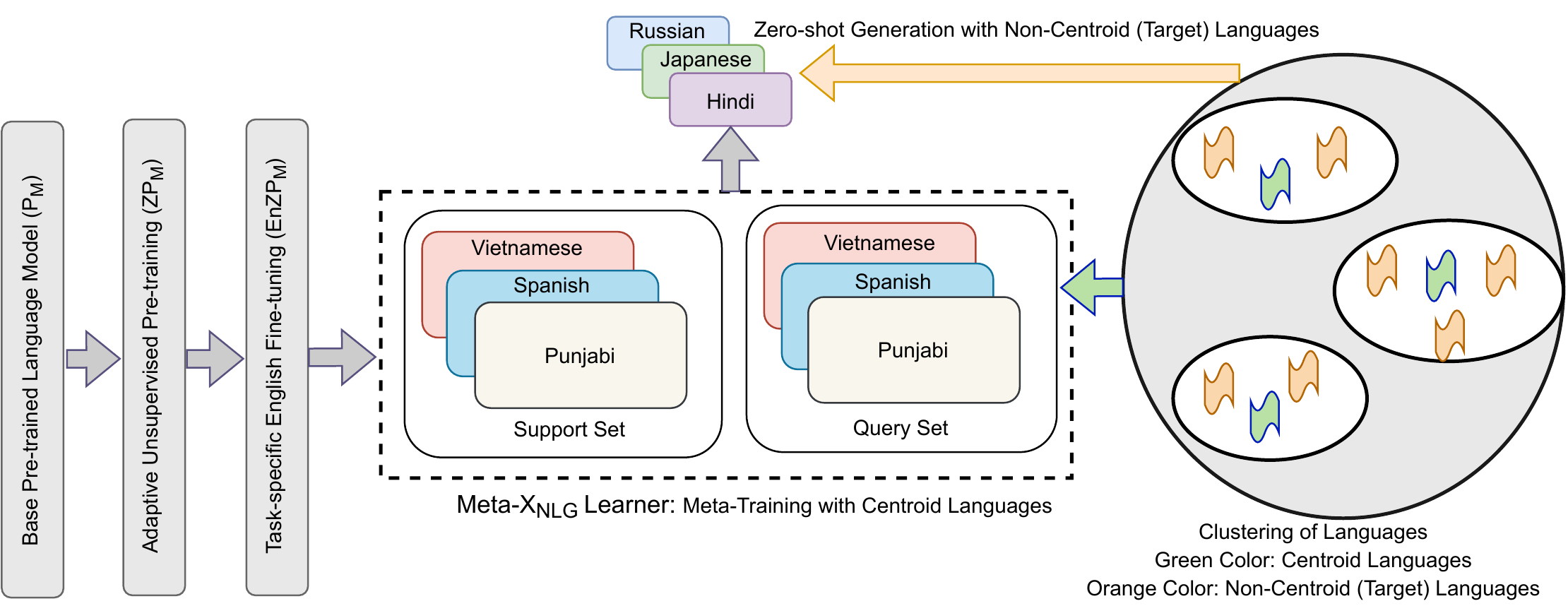}
    \vspace{-.12in}
    \caption{\small An overview of Meta-X\textsubscript{XNLG} framework}
    \label{fig:ov_meta_xnlg}
    \vspace{-.10in}
\end{figure*}

In the proposed Meta-X\textsubscript{NLG} framework, we first cluster the available languages and identify the centroid languages. Then we train a model with MAML on centroid languages to obtain an optimal initialization of parameters. Finally, the learned model is deployed to generate text in the zero-shot setting. 
Figure-\ref{fig:ov_meta_xnlg} provides an overview of proposed framework. We now provide details of each component of the framework.

\subsection{Language Clustering}
\label{sec:langc}
Broadly, the languages can be clustered in two ways: (1) By language family consideration and (2) By exploiting similarities among learned language representations. To learn language representations, \citet{littell-etal-2017-uriel} used typological information from linguistic knowledge-bases like WALS \cite{dryer2013world} Glottolog \cite{hammarstrom2017glottolog}, etc. \citet{malaviya-etal-2017-learning} extract learned language tag representations from tasks like machine translation. Recently, \citet{oncevay-etal-2020-bridging} fuse typologically learned and task-learned language representations using singular vector canonical correlation (SVCC) analysis to obtain multi-view language representation. Further, the authors cluster languages using this rich multi-view language representations through hierarchical clustering. We utilize this clustering approach in our proposed framework. 

Next, we aim to identify a representative language (\textit{centroid language}) for each cluster. Formally, given a cluster $C = \{L_1, L_2, \dots L_t \}$, where each $L_i$ is multi-view representation of $i^{th}$ language, the centroid language $L^*\in C$ is defined as:
\begin{equation}
    L^* = \arg\min_{L_i \in C} \sum_{L_j \in C} d(L_j, L_i).
\end{equation}

We use $d$ as the cosine distance. In the proposed meta-learning algorithm, the centroid languages act as \textit{Meta-Training} tasks/languages and the rest of the non-centroid languages across clusters act as \textit{Target} (aka. evaluation) tasks/languages. In this setup, the best performing model should hold two properties i.e., \textit{Intra-cluster Generalization} and \textit{Inter-cluster Generalization}. In the proposed framework, training with a centroid language leads to better transfer capability within cluster, and using multiple centroid languages extend the transfer capability to multiple closely-knit clusters and increase coverage. In this way the stated properties can be achieved. 

However, there is a trade-off between the number of clusters (the number of meta-training languages) and generalization. 
If there is a single cluster (a single meta-training language), then the model tries to over-generalize for different typological structures and fails in the attempt. On the other extreme, if there are too many centroid languages (many typologically diverse structures in meta-training), then the learning possibly gets distracted. In both cases, the model will be unable to learn a reasonable structure (the required generalization) and perform poorly. Section-\ref{sec:train_lang_effect} consists discussions and empirical evidence. Our experiments suggest that three clusters across considered languages provide the best performance. These three clusters are always fixed irrespective of the datasets and underlying tasks. Composition of the clusters (with three clusters) are shown in Table-\ref{tab:lan_cluster}. See Figure-\ref{fig:meta_xnlg_lang_cluster} for more details on the clustering. 

\begin{table}[!b]
\centering
\scalebox{0.8}{
\begin{tabular}{|c|c|c|}
\hline
\textbf{Cluster-1(14)} & \textbf{Cluster-2(8)} & \textbf{Cluster-3(8)} \\
 \hline
 hi,ur,te,tr,ja,fi,ko,gu, & es,it,pt,ro,  & ru,cs,vi,th, \\
 bn,mr,np,ta,pa,sw & nl,de,en,fr & zh,id,el,ar \\
\hline 
\end{tabular} 
}
\vspace{-.05in}
\caption{\small Clustering of considered 30 Languages}
\label{tab:lan_cluster}
\vspace{-.15in}
\end{table}

\subsection{Meta-X\textsubscript{NLG} Training}
\label{metaxnlg}

The framework consists of five training steps: Selection of Base Pre-trained model, Adaptive unsupervised pre-training, Fine-tuning with HRL, Meta-training with LRLs, and Meta-adaptation for Zero-shot. The motivation and details of each step are included below: 

\begin{enumerate}[leftmargin=*]
    \itemsep-0.12em 
    \item \textit{Selection of Base Pre-trained Model ($P_M$):} Our approach is model-agnostic, therefore any state-of-the-art sequence-to-sequence multilingual pre-trained language model ($P_M$; like mBART, mT5, etc.) can be used. We selected mT5  due to its superiority on many NLP tasks \cite{xue-etal-2021-mt5}. 
    
     \item \textit{Adaptive Unsupervised Pre-training ($ZP_M$):} Zero-shot cross-lingual generation often suffers from accidental translation \cite{xue-etal-2021-mt5} and other generation problems. To overcome this, we \textit{further train} $P_M$ on a \textit{MultiMonoLang} corpus with mT5 denoising objective. We created \textit{MultiMonoLang} corpus by concatenating small unsupervised samples from each of the 30 languages. We call this model $ZP_M$ (or ZmT5). See section-\ref{subsec:at_cf} for more details. 
     
     \item \textit{Fine-tuning $ZP_M$ on High Resource Language (i.e., English):} It is often observed that downstream LRLs applications benefit when supervision is transferred from HRL \cite{hu2020xtreme}. Following the trend, we fine-tune the $ZP_M$ model with the task-specific English data and call this model as $EnZP_M$ with parameters $\theta_p$. 
    
     \item \textit{Meta-Training with Low-resource Centroid Languages:} We use the validation sets of each centroid language as the \textit{meta-train} dataset. The meta-learner is initialized with the $EnZP_M$ parameters. Then, a batch of tasks/languages $T_i$ and corresponding datasets $D_i$ are randomly sampled. Further, each $D_i$ is equally split into support set $S_i$ and query set $Q_i$ such that they are mutually exclusive. $m$-step gradient update is done in the inner-loop using $S_i$. This is repeated for all the training tasks. Finally\textit{ Meta-Update} is done using mean loss computed on $Q_i$ as shown in Equation 2. This is repeated for all the tasks/languages over multiple batches. The batches are sampled uniformly across all centroid languages. The formal description is shown in Algorithm-\ref{mxmlg_algo}.
     
     \item \textit{Meta-adaptation for Zero-shot Evaluation:} The meta-learned model $f_{\theta^*}$ from the previous step can be directly evaluated on the test sets of the target languages in zero-shot evaluation. The proposed framework can be easily extended to few-shot setting. In this setting, the meta-learned model can be fine-tuned on a small number of validation set examples with standard supervised learning and evaluated on the test sets of target languages. In this work we consider zero-shot setting only.    
\end{enumerate}

\begin{algorithm}
\small
\caption{Meta Learning Algorithm}
\begin{algorithmic}[1]
\Require Task set distribution $p(D)$, pre-trained model $EnZP_M$ (P) with parameters $\theta_P$, meta-learner $f_{\theta}$ with parameter $\theta$.
\Require $\alpha$, $\beta$: step size hyper-parameters 

\State Initialize $ \theta \gets \theta_P $
\While {not done}
    \State Sample batch of tasks $T={T_1, T_2, \dots T_b} \sim p(D)$ 
    \For{all $T_i$ in $T$ }
    \State Initialize $\theta_i \gets \theta$
    \State Split $ D_i$ to form support set $S_i$ and query set $Q_i$ 
        \For{ all inner\_iter steps  $m$}
        \State Compute $\nabla_{{\theta}_i^{(m)}} L_{T_i}^{S_i}(P_{{\theta}_i^{(m)}})$
        \State Do SGD:  $ \theta_i^{m+1}$ = $\theta_i^{m} -\alpha \nabla_{{\theta}_i^{(m)}} L_{T_i}^{S_i}(P_{{\theta}_i^{(m)}})$
        \EndFor
        \State MetaUpdate: $\theta = \theta - \beta \nabla_{\theta} \sum_{j=1}^b L_{T_i}^{Q_i}(P_{{\theta}_i^{(m)}})$
    \EndFor
\EndWhile
\State Do zero-shot/few-shot learning with meta-learner $f_{\theta^*}$ where $\theta^*$ is learned optimal parameters of meta-learner.
\end{algorithmic}
\label{mxmlg_algo}
\end{algorithm}

\subsection{Avoiding Accidental Translations:}
\label{subsec:at_cf}
\vspace{-.07in}
It has been observed that popular pre-trained models like mBART and mT5 suffer in well-formed generation for unseen low-resource (zero-shot) languages. Broadly, they suffer from Accidental Translation (AT), where the model generates whole/part of the output in the fine-tuning language  \cite{xue-etal-2021-mt5}. This happens when the model forgets the learning obtained before fine-tuning. This is analogous to the Catastrophic-Forgetting problem \cite{chi2020cross} in multi-task setup, where the model forgets the learnings about the previous task. 
For language generation, this also leads to problems like improper predictions, structural and normalization errors, etc., as the different languages differ in morphology, phonology, subject-verb-object ordering, etc. 

To mitigate/reduce these problems, \citet{xue-etal-2021-mt5} suggested mixing a small amount of multilingual pre-training task data into the fine-tuning stage. However, it is unclear what ratio mixing should be done and how this joint training will affect generation quality. Moreover, such mixing is not a feasible solution for multi-level fine-tuning (as in our proposed setup - English fine-tuning then meta-training with centroid languages). Inspired from \citet{DBLP:conf/acl/MauryaDKD21}, the following solution approach are adopted in Meta-X\textsubscript{NLG} framework.
\begin{itemize}[leftmargin=*]
   \itemsep-0.2em 
    \item \textit{Adding Language Tag:}  We concatenate \textit{<fxx> <2xx>} where \textit{xx}
    is language code as per ISO 693-2 standard. 
    \item \textit{Adaptive Unsupervised Pre-training}: Further train the base pre-trained model on \textit{MultiMonoLang} corpus with denoising language model objective. Unlike \citet{DBLP:conf/acl/MauryaDKD21}, we use mT5 denoising objective \cite{xue-etal-2021-mt5} instead \textit{rand-summary} objective which leads to better performance.  
    \item \textit{Freezing model Components :} One of the key approaches to mitigate CF problem is freezing model parameters. \citet{DBLP:conf/acl/MauryaDKD21} performed an ablation study and concluded that freezing all token embeddings and decoder parameters of the model work best. We adapted these findings while English-fine tuning and meta-training steps.
\end{itemize}
We observed that the above settings work better to mitigate (or reduce) the AT problem. See Table-\ref{tab:adpobj} in appendix for ablation study results. 

\section{Experiment Setup}
\label{sec:exp_setup}

We investigate Meta-X\textsubscript{NLG}'s performance on two NLG tasks, five datasets and 30 languages. mT5 pre-trained model is used as the base model. The model performance is compared with two strong baselines in zero-shot setting.  
\subsection{Tasks and Datasets}
\label{sec:task_data}
\subsubsection{Abstractive Text Summarization (ATS):} ATS is the task of \textit{generating grammatically coherent, semantically correct and abstractive summary given an input document}. We use two publicly  available datasets: XL-Sum \cite{hasan-etal-2021-xl} and Wikilingua \cite{ladhak-etal-2020-wikilingua}. \\
\textbf{XL-Sum} is a large comprehensive dataset where article-summary pairs are extracted from BBC and annotated by professional annotators. It covers 44 languages including very low-resource languages like Nepali and Swahili. Due to computational limitation, we consider only 23 languages. \\
\textbf{Wikilingua} is a large-scale dataset covering 18 languages. Article and summary pairs are extracted from WikiHow\footnote{\url{https://www.wikihow.com/}}. It is \textit{how-to guides} on diverse topics written by human annotators. We consider all 18 languages in our experiments.

\subsubsection{Question Generation (QG):} In QG, \textit{given an input passage and an answer, it aims  to generate semantically and syntactically correct questions that can produce the answer}. We use three publicly available multilingual question and answering (QA) datasets:  MLQA \cite{lewis-etal-2020-mlqa}, TyDiQA \cite{clark-etal-2020-tydi} and XQuAD \cite{artetxe-etal-2020-cross}. Each instance is triplet of <passage, question, answer>. We concatenated \textit{answer} and \textit{passage} with delimiter \textit{</s>} in same order as input for models.

\textbf{MLQA} is a multi-way parallel extractive QA evaluation dataset available in 7 languages. Authors  automatically extracted  paragraphs from Wikipedia articles in multiple languages which have same or similar meaning. Authors crowd-source questions on English and translate into target languages by professional translators. As our frame-work is based on supervision transfer we only consider the evaluation data instance where input and target text languages are same. In this way we have 7 datasets for 7 languages. 

\textbf{XQuAD} dataset is translated from the development set of SQuAD v1.1 \cite{rajpurkar-etal-2016-squad} by professional human translators into 10 languages. Each languages has 1190 question-answer pairs.  SQuAD is popular  question answering dataset consisting of around 100k <passage, question, answer> triplets. 
We added additional Japanese language data set \cite{takahashi-etal-2019-machine} which is created with similar goals and has same format.

\textbf{TyDiQA} is another QA dataset with 204K question-answer pairs in 11 typologically diverse languages. Unlike MLQA and XQuAD, it is directly collected in each language and does not involve any translation. We use, \textit{TyDiQA-GoldP} datasets which is guaranteed to have extractive nature. We added Tamil as additional language that share same format and created with similar goals.

We use English data from XL-Sum and Wikilingua for English fine-tuning step while experimenting with respective dataset. MLQA, TyDiQA and XQuAD  do not have any English training data. Following the trend \cite{lewis-etal-2020-mlqa, clark-etal-2020-tydi} we use SQuAD v1.1 training data at English fine-tuning step.

For each dataset, we grouped the languages into three fixed clusters as per Table-\ref{tab:lan_cluster} and find the centroid language as described in  Section-\ref{sec:langc}. English is the high resource language and only used for supervised fine-tuning as described in section-\ref{metaxnlg} so, it will not be part of any cluster. To make it more concrete, XQuAD dataset has 11 low-resource languages (excluding English), the centroid (Meta-training) languages are <tr,es,th> and non-centroid (Target) languages are <hi,ro,de,ar,vi,zh,ru,el>\footnote{see Table-\ref{tab:lang_cluster} for language distribution to the cluster for each dataset and Table-\ref{tab:data_stats} for datasets statistics}. 

\subsection{Baselines}
Due to unavailability of prior zero-shot results for considered datasets, we design strong baselines based on recent model architectures. 
\begin{itemize}[leftmargin=*]
   \itemsep-0.3em 
    \item \textbf{EnZmT5:} Inspired from \citet{DBLP:conf/acl/MauryaDKD21}, we further train mT5 model with monolingual dataset in all 30 languages followed by task-specific English fine-tuning (similar to first three steps of Meta-X\textsubscript{NLG} model proposed in section -\ref{metaxnlg}). Then it is directly evaluated on the target languages in zero-shot setting. 
    \item \textbf{FTZmT5:} In this model we fine-tune EnZmT5 baseline on all centroid languages. This will ascertain that the improvement of Meta-X\textsubscript{NLG} is not due to simply training on more datasets in different languages. This is close to the \citet{lewis2020pretraining}'s model but they use different dataset.
\end{itemize}

While training EnZmT5 and FTZmT5, we use all applicable precautions as suggested in sections-\ref{subsec:at_cf} and grid search to find best hyper-parameters. We could not compare ZmBART performance with Meta-X\textsubscript{NLG} as authors did not use officially released evaluation datasets\footnote{for Wikilingua dataset, official splits are released recently.}. 

\subsection{Evaluation Metrics}
\label{sec:eval_met}
Both automatic and manual evaluation metrics are used to ensure the quality of the generated text. Particularly, for automatic evaluation \textbf{ROUGE-L} \cite{lin-2004-rouge} and \textbf{BLEU}\footnote{reported scores are $\tt{case-mix}$ BLEU-4 from modified sacreBLEU implementation, see appendix-\ref{sec:app_humEval}} \cite{10.3115/1073083.1073135} metrics are used for ATS and QG respectively. 
Similar to \citet{Chi_Dong_Wei_Wang_Mao_Huang_2020} we used three manual evaluation metrics: \textbf{Fluency} referring to \textit{how fluent the generated text is}, \textbf{Relatedness} indicating \textit{the degree of the input's context in the generated text} and  \textbf{Correctness} measuring the \textit{grammar and semantics of generated text}. It is often observed that NLG systems suffer from the problem of Hallucination \cite{Nie2019ASR}; the \textit{Relatedness} metric provides clarity in such situations. The \textit{Correctness} metric is hard metric which considers both semantic and grammatical aspects of generated text. 

We randomly sampled 50 generated examples for each <task, dataset, language> triplet based on qualified and available native language experts in Hindi, Telugu, Tamil and Bengali languages. In total, we selected six triplets for evaluation. To ensure the quality, each selected triplet is evaluated by two sets of annotators. We asked each annotator to rate the generated text on a scale of 1-5 (where 1 is very bad and 5 is very good) for the metrics mentioned above. We anonymously shared the generated text from two baselines and Meta-X\textsubscript{XNLG} to avoid any biased evaluation.
\subsection{Implementation Details}
\label{sec:app_imp}
We implemented Meta-X\textsubscript{NLG} using \textit{higher} library\footnote{\url{https://github.com/facebookresearch/higher}}. SGD with learning rate ($\alpha$) $1e-4$ is used as inner-loop optimizer and AdamW with learning rate ($\beta$) $1e-5$ is used as outer-loop optimizer. The inner iteration (\textit{m}) value is 2 and  meta-training batch size is 8. To partition the training batch into support set ($S$) and query set ($Q$), we experimented (S:Q) with [8:2, 7:3, 6:4, 5:5, 4:6] splits. The best results are obtained with equal partition, i.e., 5:5. We also experimented with [2, 5, 10, 15, 20, 25] training epochs. The best performance was observed at $10^{th}$ epoch. We use a standard mT5-small sequence-to-sequence Transformer architecture with 12 layers (each 16 heads). It has 1024 dimensions and approx 582M parameters. Additional layer-normalization with weight decay (0.1) was used with both the encoder and decoder. For input, the max sequence length is fixed to 512. We trained all the models on 1
Nvidia V100 GPU (32GB). Cross-entropy label smoothing is used as loss function. We use beam-search
with beam size 4; max generation length is 100 for ATS (32 for QG) and min length is 1. To ensure the stated improvement on the MLQA dataset, we compute average BLEU scores across the best 5 checkpoints. We are unable to repeat such experiments for other datasets due to computational limits.


\begin{table*}[!htb]
\centering
\scalebox{0.6}{
\begin{tabular}{l|c|c|c|c|c|c|c|c|c|c|c|c|c|c|c|c|c|c|c}
    \hline\hline
 \textbf{Model} & \textbf{fr} &  \textbf{gu}  &  \textbf{id}  &   \textbf{th}  &   \textbf{ta}  &  \textbf{hi} & \textbf{mr} & \textbf{ja} & \textbf{ko} & \textbf{tr} & \textbf{ru} & \textbf{sw} & \textbf{pt} & \textbf{ar} & \textbf{te} & \textbf{ur} & \textbf{ne} & \textbf{bn} & \textbf{zh}      \\
 \hline
\textbf{EnZmT5} & 18.45 & 13.21 & 19.77 & 21.53 & 11.58 & 22.24 & 11.89 & 22.81 & 18.74 & 17.72 & 15.27 & 18.91 & 18.92 & 18.44 & 10.77 & 21.61 & \textbf{16.24} & 16.12 & 21.07\\
\textbf{FTZmT5} & 21.83 & 7.98 & 19.27 &\textbf{ 24.68} & 10.80 & 11.92 & 8.94 & 23.32 & 16.82 & 14.99 & 12.90 & 21.01 & 20.07 & 15.85 & 9.14 & 13.05 & 11.06 & 12.66 & 15.20 \\
\textbf{Meta-X\textsubscript{NLG}}& \textbf{22.83} & \textbf{14.02} & \textbf{21.54} & 24.61 & \textbf{12.88} & \textbf{23.09} & \textbf{12.58} & \textbf{25.33} & \textbf{20.12} & \textbf{18.65} & \textbf{17.31} & \textbf{22.63}  & \textbf{20.24} & \textbf{20.11} & \textbf{12.07} & \textbf{23.41} & 15.45 & \textbf{17.96} & \textbf{22.95}\\
\hline \hline
\end{tabular} 
}
\caption{\small Zero-shot \textit{Rouge-L} scores for 19 target languages on XL-Sum dataset \cite{hasan-etal-2021-xl}. EnZmT5 \cite{DBLP:conf/acl/MauryaDKD21} and FTZmT5 are baseline models. Scores are reported after extensive hyper-parameter search for all the models.}
\label{tab: rslt_xlsum}
\end{table*}

\begin{table*}[!htb]
\centering
\scalebox{0.75}{
\begin{tabular}{l|c|c|c|c|c|c|c|c|c|c|c|c|c|c}
    \hline\hline
 \textbf{Model} & \textbf{id} &  \textbf{fr}  &  \textbf{ar}  &   \textbf{pt}  &   \textbf{it}  &  \textbf{th} & \textbf{ru} & \textbf{cs} & \textbf{nl} & \textbf{de} & \textbf{ja} & \textbf{zh} & \textbf{hi} & \textbf{tr}\\
 \hline
\textbf{EnZmT5} & 15.34 & 18.72 & \textbf{15.70} & 17.21 & 15.05 & 26.66 & 14.67 & 9.42 & 17.97 & 13.69 & 22.32 & 20.12 & 18.88 & 14.45\\
\textbf{FTZmT5} & 13.69 & 19.37 & 12.66 & 17.80 & 15.54 & 23.72 & 11.95 & 10.20 & 16.74 & 12.22 & 22.81 & 18.64 & 17.32 & 13.84\\
\textbf{Meta-X\textsubscript{NLG}} & \textbf{16.85} &\textbf{ 20.26} & 15.66 & \textbf{18.36} & \textbf{16.03} & \textbf{27.71} & \textbf{14.89} & \textbf{11.76} & \textbf{19.09} & \textbf{14.11} & \textbf{22.83} &\textbf{ 22.45} & \textbf{19.60} & \textbf{15.23}\\
\hline \hline
\end{tabular} 
}
\caption{\small Zero-shot \textit{Rouge-L} scores for 14 target languages on Wikilingua   dataset \cite{ladhak-etal-2020-wikilingua}.}
\label{tab: rslt_xlsum}
\end{table*}

\begin{table*}[!htb]
\centering
\scalebox{0.9}{
\begin{tabular}{l|c|c|c|c|c|c|c|c}
    \hline\hline
 \textbf{Model} & \textbf{ar} &  \textbf{de}  &  \textbf{zh}  &   \textbf{vi}  &   \textbf{hi}  &  \textbf{el} & \textbf{ru} & \textbf{ro} \\
 \hline
\textbf{EnZmT5} & 8.55 & 9.99 & 23.76 & 17.29 & 9.55 & 8.18 & 10.98 & 11.27\\
\textbf{FTZmT5} & 5.82 & 9.040 & 22.87 & 16.47 & 9.05 & 6.95 & 8.87 & 10.31 \\
\textbf{Meta-X\textsubscript{NLG}} & \textbf{8.63} & \textbf{10.52} & \textbf{24.89} & \textbf{20.92} & \textbf{11.90} & \textbf{9.01} & \textbf{11.41} & \textbf{12.24}\\
\hline \hline
\end{tabular} 
}
\caption{\small Zero-shot \textit{BLEU} scores for 8 target languages on XQuAD  dataset \cite{artetxe-etal-2020-cross}.}
\label{tab: rslt_xquad}
\end{table*}

\begin{table*}[!htb]
\parbox{.40\linewidth}{
\centering
\scalebox{0.85}{
\begin{tabular}{l|c|c|c|c|c|c|c}
    \hline\hline
 \textbf{Model} & \textbf{fi} &  \textbf{ru}  &  \textbf{id}  &   \textbf{sw}  &   \textbf{ko}  &  \textbf{bn} & \textbf{ta} \\
 \hline
\textbf{EnZmT5} & 7.87 & 5.52 & 5.75 & 4.48 & 8.59 & 5.77 & 3.08\\
\textbf{FTZmT5} & 8.39 & 7.28 & \textbf{11.42} & 5.51 & 10.05 & 7.96 & 2.022\\
\textbf{Meta-X\textsubscript{NLG}} & \textbf{9.08} & \textbf{7.47} & 9.36 & \textbf{6.42} &\textbf{ 12.67} & \textbf{9.17} & \textbf{9.76}\\
\hline \hline
\end{tabular} 
}
\caption{\small Zero-shot \textit{BLEU} scores on TyDiQA  data.}
\label{tab: rslt_tydiqa}
}
\hfill
\parbox{.40\linewidth}{
\centering
\scalebox{0.85}{
\begin{tabular}{l|c|c|c|c}
    \hline\hline
 \textbf{Model} & \textbf{hi} &  \textbf{es}  &  \textbf{ar}  &   \textbf{zh}  \\
 \hline
\textbf{EnZmT5} & 5.06 & 6.94 & 3.46 & 13.70\\
\textbf{FTZmT5} & 5.14 & 6.16 & 2.21 & 13.38\\
\textbf{Meta-X\textsubscript{NLG}} & \textbf{5.66} & \textbf{7.03} & \textbf{3.66} & \textbf{15.13} \\
\hline \hline
\end{tabular} 
}
\caption{\small Zero-shot \textit{BLEU} scores on MLQA  data.}
\label{tab:rslt_mlqa}
}
\end{table*}

\begin{table*}[!htb]
\centering
\scalebox{0.9}{
\begin{tabular}{l|l|c|c|c||l|c|c|c}
    \hline\hline
\textbf{Model} & \textbf{Task/Data/Lang} & \textbf{Flu} &  \textbf{Rel}  &  \textbf{Corr}  &   \textbf{Task/Data/Lang} & \textbf{Flu} &  \textbf{Rel}  &  \textbf{Corr}  \\
 \hline
\multicolumn{5}{l}{\textit{Annotator set-1}} \\\hline
\textbf{EnZmT5} & &4.06 & 3.58 & 2.84 & & 4.28 & 3.94 & 3.70 \\
\textbf{FTZmT5} & ATS/XL-Sum/bn & 2.82 & 3.18 & 2.08 & ATS/XL-Sum/te & 3.46 & 3.46 & 3.22\\
\textbf{Meta-X\textsubscript{NLG}} & &\textbf{4.12}  & \textbf{4.34} & \textbf{3.44} & & \textbf{4.50} & \textbf{4.22} & \textbf{4.04} \\
\hline 
\multicolumn{5}{l}{\textit{Annotator set-2}} \\\hline
\textbf{EnZmT5} &  & 3.70 & 3.23 & 3.26 & & 3.56 & 3.50 & 3.20\\
\textbf{FTZmT5 }& ATS/XL-Sum/bn & 2.62 & 2.48 & 2.16 & ATS/XL-Sum/te & 3.02 & 2.84 & 2.60\\
Meta-X\textsubscript{NLG} & & \textbf{3.97} & \textbf{3.48} & \textbf{3.28} & & \textbf{4.18} & \textbf{4.10} & \textbf{3.88} \\
\hline \hline
\multicolumn{5}{l}{\textit{Annotator set-1}} \\\hline
\textbf{EnZmT5} &  & 4.00 & 3.72 & 3.68 & & 4.12 & 4.24 & 2.54 \\
\textbf{FTZmT5} & ATS/Wiki/hi & 4.07  & 3.39 & 3.83 & QG/XQuAD/hi & 4.22 & 4.02 & 2.56\\
\textbf{Meta-X\textsubscript{NLG}} &  & \textbf{4.09} & \textbf{3.80} & \textbf{3.97}& & \textbf{4.42} & \textbf{4.34} & \textbf{2.86}\\
\hline 
\multicolumn{5}{l}{\textit{Annotator set-2}} \\\hline
\textbf{EnZmT5} &  & 4.38 & 4.22 & 4.00 & & 3.28 & 3.63 & 2.82\\
\textbf{FTZmT5 }& ATS/Wiki/hi & 4.57 & \textbf{4.44} & 4.08 & QG/XQuAD/hi & 3.24 & 3.34 & 2.89\\
\textbf{Meta-X\textsubscript{NLG}} &  & \textbf{4.66} & \textbf{4.44} & \textbf{4.16} & & \textbf{3.59}& \textbf{3.67} & \textbf{3.24}\\
\hline \hline
\multicolumn{5}{l}{\textit{Annotator set-1}} \\\hline
\textbf{EnZmT5} &  & 3.48 & 3.70 &3.46 & & 4.25 & 4.06 & 3.10\\
\textbf{FTZmT5} & QG/MLQA/hi &  3.44 & 3.42 & 3.18 & QG/TyDiQA/ta & 3.25& 3.01 & 2.07\\
\textbf{Meta-X\textsubscript{NLG}} &  & \textbf{3.70} & \textbf{3.74} & \textbf{3.56} & &\textbf{4.74} & \textbf{4.20} & \textbf{3.39}\\
\hline 
\multicolumn{5}{l}{\textit{Annotator set-2}} \\\hline
\textbf{EnZmT5} &  & \textbf{3.30} & 3.28 & 2.40 & & 3.00 & 4.08 & 2.82\\
\textbf{FTZmT5} & QG/MLQA/hi & 3.10 & 3.44 & 2.84 & QG/TyDiQA/ta & 2.55 & 3.045 & 1.83 \\
\textbf{Meta-X\textsubscript{NLG}} & & 3.24 & \textbf{3.7}0 & \textbf{2.88} &  & \textbf{4.04} & \textbf{4.46 }& \textbf{3.20}\\
\hline \hline
\end{tabular} 
}
\caption{\small Human Evaluation results for four languages (\textbf{hi: }Hindi, \textbf{te: }Telugu, \textbf{ta: }Tamil and \textbf{bn: }Bengali), two annotator sets, two tasks (\textbf{ATS} and \textbf{QG}) and all five datasets. \textbf{Flu:} Fluency, \textbf{Rel:} Relatedness and \textbf{Corr:} Correctness metrics. Results are shown for two annotation sets which ensure biased free evaluation. Reported scores are average of all the annotators in a annotator set.}
\label{tab: ruslt_humEval}
\end{table*}


\section{Results and Analysis}
\label{sec:resluts}
Automated evaluation results are shown in Table \ref{tab: rslt_xlsum}-\ref{tab:rslt_mlqa}. \mxnlg~ consistently outperformed other two baselines on all five datasets and most of the languages. For the summarization task, among the 33 experiments (19 languages for XL-Sum and 14 for Wikilingua) \mxnlg~ gives best performance for 30 experiments. Wherever it loses out, it does so by small margin. We see that the performance gains for the Wikilingua are relatively smaller. This might be due to the nature of the Wikilingua dataset,  we observe that the input documents are set of usage instructions for softwares/tools. For such data, many instructions need to be retained in the summary. This poses a challenge to all the models including \mxnlg. Similar observations are made by \citet{DBLP:conf/acl/MauryaDKD21}.


For the question generation task, \mxnlg~ achieves better performance than others except for one experiment - Indonesian language for TyDiQA. For MLQA, improvements achieved by the proposed model are marginal (see Table-\ref{tab:rslt_mlqa}). Upon close inspection, we notice that MLQA had small number of languages, and the centroid languages are very distinct, i.e. they have higher mean distance to other languages from same cluster as compared to the other datasets (see Table-\ref{tab:lang_cluster}). This might be a possible reason for such performance.


The human evaluation scores for all the three metrics are shown in Table-\ref{tab: ruslt_humEval}. The human evaluations (across both annotator sets) correlate with automatic evaluations. Similar to the automatic evaluation, Meta-X\textsubscript{NLG} consistently outperformed both baselines for selected languages, tasks and datasets.  
High \textit{Fluency} and \textit{Relatedness} scores for Meta-X\textsubscript{NLG} indicates that most of generated text are fluent and not hallucinated respectively. The correctness metric considers both semantic and grammatical aspects; good scores on this metric indicate the acceptable performance for the proposed model in zero-shot setting. In QG, generating well-formed interrogative sentences is challenging, particularly in zero-shot setting due to unseen interrogative syntax structure of target language \cite{mitra2021zero, DBLP:conf/acl/MauryaDKD21}. The above-average fluency and correctness score for Meta-X\textsubscript{NLG} indicates that the model quickly adapts such syntax and performs better. 

The consistent improvement in Meta-X\textsubscript{NLG} for most the typologically diverse target languages provides evidence that supervision transfer is more uniform. Considering decent automatic and manual evaluation scores in the zero-shot setting, we conclude that our model performs reasonably well except small performance gain with the MLQA dataset. 
Meta-X\textsubscript{NLG} is a zero-shot framework, and we do not assume any prior training/knowledge for new unseen LRL. The only constraints are:  the new language should be part of base pre-trained models (mT5) and adaptive unsupervised pre-training (uses task-agnostic monolingual data only). Hence, adding new languages in Meta-X\textsubscript{NLG} is a simple extension exercise.
\subsection{Cross-lingual Transfer:}
\begin{figure}[!htb]
    \centering
    \subfloat[\centering Baseline]{{\includegraphics[width=3.3cm]{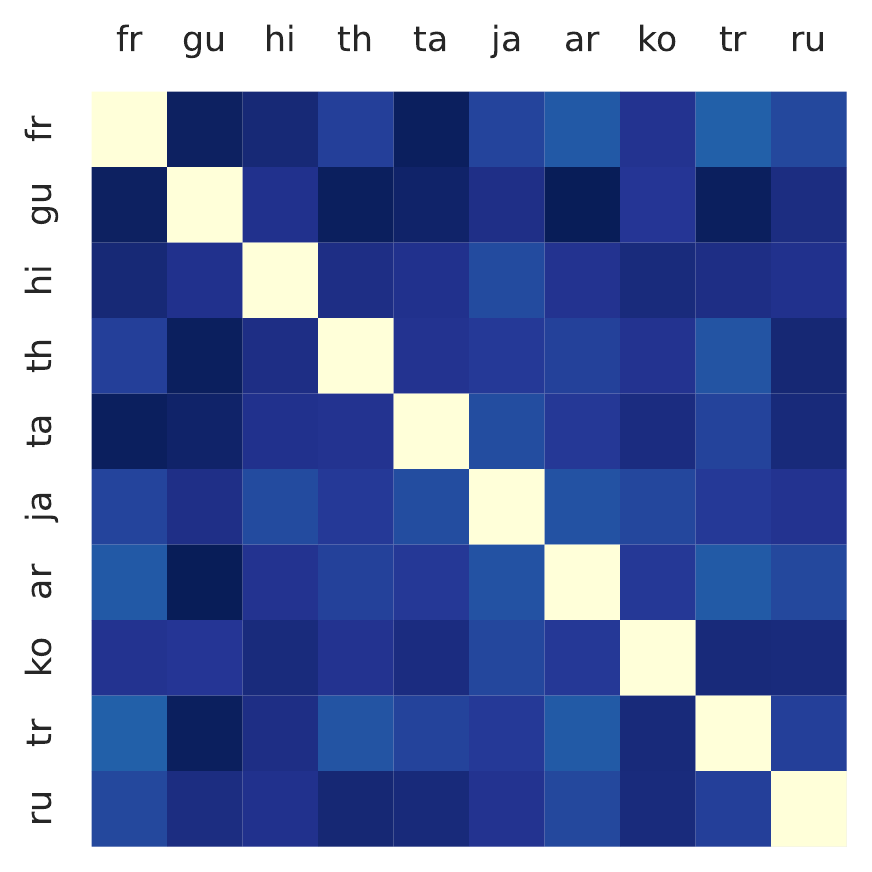} }}%
    \qquad
    \subfloat[\centering Meta-X\textsubscript{NLG}]{{\includegraphics[width=3.3cm]{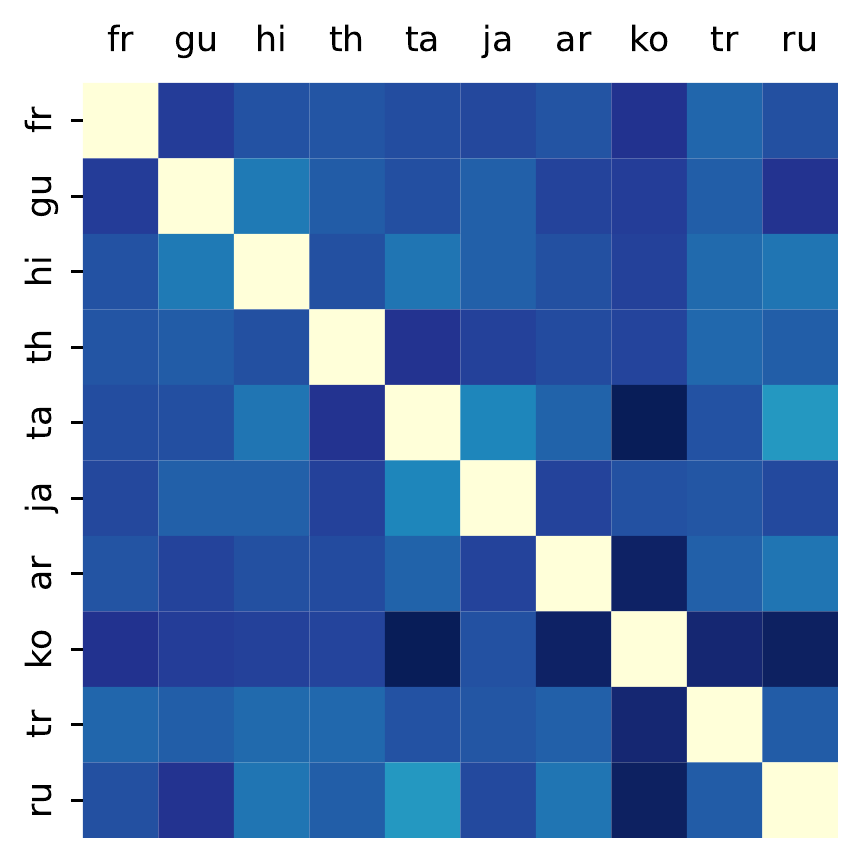} }}%
    \vspace{-.12in}
    \caption{\small Cosine distance between language tags obtained from EnZmT5 and Meta-X\textsubscript{NLG} for 10 languges from XL-Sum dataset. Dark color indicate higher cosine distance.}
    \label{fig:xlingtrans}
\end{figure}

To have a more general view of the model's learning of multiple languages, we perform similarity analysis among representations of the language tags (contextual representation of the \textit{<fxx> <2xx>} tokens from the beginning of the input in language \textit{xx}). 10 languages are randomly selected from XL-Sum dataset. Each language input is passed through the encoder part of the models (EnZmT5 and \mxnlg) and language tag representations (LTRs) are extracted. Cosine distance among LTRs is shown in figure-\ref{fig:xlingtrans}. Baseline EnZmT5 has a high cosine distance between LTRs and the shared latent representation space is not much aligned. Meta-X\textsubscript{NLG} has lower distances and shared latent representation space is more aligned across languages. 

\subsection{Effect of Training Languages:}
\label{sec:train_lang_effect}
Table-\ref{tab: rslt_anays1} shows the results with different language combinations for Meta-X\textsubscript{NLG} training on XQuAD dataset. For this dataset, the centroid languages are Turkish (tr), Spanish (es) and Thai (th). Results are generally good when centroid languages are in the training set. Best results are obtained using three centroid languages from three clusters. The performance dropped when we included more centroid languages (rows 12-15). As discussed in section-\ref{sec:langc}, learning gets distracted with many centroid languages.

We now try to have a closer look at the numbers. While training with non-centroid languages (rows 4, 8, 9), the model performs poorly, which validates the importance of centroid languages. Another example is \textit{Turkish} and \textit{Hindi} languages share same cluster, in row 5 we did not include \textit{Turkish} as centroid language which obtains poor performance on \textit{Hindi}. Similar observations can be made for row-6. Overall, Meta-X\textsubscript{NLG} trained with three centroid languages (row 14) performs best on most of the languages and on average. 
We conducted more extensive ablation study with XL-Sum dataset (see Table-\ref{tab:xlsum_rslt_anays1} in Appendix) and similar trends are observed.    

\begin{table}
\centering
\scalebox{0.5}{
\begin{tabular}{c|l|c|c|c|c|c|c|c|c|c}
    \hline\hline
\textbf{SetUp} & \textbf{MTrain Lang} & \textbf{ar} &  \textbf{de}  &  \textbf{zh}  &   \textbf{vi}  &   \textbf{hi}  &  \textbf{el} & \textbf{ru} & \textbf{ro} & \textbf{avg}\\
 \hline
 1 & tr & 6.14 & 8.61 & 23.67 & 19.81 & \textbf{10.91} & 6.80 & 9.53 & 10.17 & 11.89 \\
 2 &es & 6.68 & 10.82 & 20.89 & 16.84 & 7.96 & 7.79 & 10.02 & 13.28 & 11.78 \\
 3 &th & 5.43 & 8.47 & 23.10 & 17.46 & 7.99 & 6.85 & 9.41 & 8.98 & 11.08\\
 4 &ro & 4.78 & 9.49 & 19.80 & 15.75 & 6.01 & - & 8.25 & 9.90 & 10.56 \\
 5 & es,th & 6.07 & 10.30 & 18.74 & 16.10 & 7.74 & 7.14 & 9.56 & 12.37 & 11.00\\
 6 & tr,th & 6.02 & 8.58 & 25.05 & 19.08 & 10.38 & 6.64 & 9.27 & 10.40 & 11.92\\
 7 & ro,de & 5.53 & - & 22.69 & 15.37 & 7.59 & 6.37 & 8.85 & - & 11.06 \\
 8 & zh,ar & - & 8.92 & - & 15.55 & 8.22 & 6.58 & 9.72 & 10.49 & 9.91\\
 9 & de,ru & 6.02 & - & 17.68 & 12.40 & 8.05 & 7.32 & - & 12.56 & 10.67\\
 10 & vi,th, el & 6.15 & 9.86 & 23.26 & - & 8.86 & - & 9.94 & 11.71 & 11.63 \\
11 & de,tr,el & 5.91 & - & 14.29 & 18.15 & 9.50 & - & 9.88 & 12.28 & 11.66 \\
 12 & tr,es,th, ru & 6.03 & \textbf{11.88} & 23.13 & 19.56 & 9.58 & 7.04 & - & \textbf{13.62 }& 12.97 \\
 13 &tr,es,th,de & 6.34 & - & 17.25 & 19.47 & 8.91 & 7.73 & 9.95 & 13.14 & 11.82 \\
 14 &tr,es,th,de,ru & 6.45 & - & \textbf{25.14} & 16.31 & 9.51 & 6.72 & - & 12.39 & 12.75  \\
 15 &tr,es,th,de,ru,ar & - & - & 22.58 & 15.65 & 8.04 & 6.74 & - & 11.81 & 12.96  \\ 
 16 &Meta-X\textsubscript{NLG} & \textbf{8.63} & 10.52 & 24.89 & \textbf{20.92} & \textbf{11.90} & \textbf{9.01} & \textbf{11.41} & 12.24 & \textbf{13.69}\\
\hline \hline
\end{tabular}
}
\caption{\small Meta-X\textsubscript{NLG} zero-shot results on different training languages combinations of the XQuAD dataset. '-' indicates the language used in training, so scores are not zero-shot and not included. 
}
\label{tab: rslt_anays1}
\end{table}

\section{Conclusion}
In this work, we propose a novel Meta-X\textsubscript{NLG} framework based on meta-learning and language clustering for effective cross-lingual transfer and generation. This is the first study that uses meta-learning for zero-shot cross-lingual transfer and generation. The evaluations are done on two challenging tasks (ATS and QG), five publicly available datasets and 30 languages. Consistent improvement for both human and automatic evaluation metrics is observed over baselines. The cross-lingual transfer analysis indicates the model's ability towards uniform cross-lingual transfer across multiple low-resource languages. We will extend this study to more cross-lingual tasks and languages in the future.

\section*{Acknowledgments}
\vspace{-.1in}
 We thank all the human annotators for human evaluation and the anonymous reviewers for their constructive feedback.

\bibliography{anthology,custom}

\begin{thebibliography}{46}
\expandafter\ifx\csname natexlab\endcsname\relax\def\natexlab#1{#1}\fi

\bibitem[{Artetxe et~al.(2020)Artetxe, Ruder, and
  Yogatama}]{artetxe-etal-2020-cross}
Mikel Artetxe, Sebastian Ruder, and Dani Yogatama. 2020.
\newblock On the cross-lingual transferability of monolingual representations.
\newblock In \emph{Proceedings of the 58th Annual Meeting of the Association
  for Computational Linguistics}, pages 4623--4637, Online. Association for
  Computational Linguistics.

\bibitem[{Ayana et~al.(2018)Ayana, Shen, Chen, Yang, Liu, and Sun}]{8370729}
Ayana, Shi-qi Shen, Yun Chen, Cheng Yang, Zhi-yuan Liu, and Mao-song Sun. 2018.
\newblock \href {https://doi.org/10.1109/TASLP.2018.2842432} {Zero-shot
  cross-lingual neural headline generation}.
\newblock \emph{IEEE/ACM Transactions on Audio, Speech, and Language
  Processing}, 26(12):2319--2327.

\bibitem[{Bansal et~al.(2020)Bansal, Jha, Munkhdalai, and
  McCallum}]{bansal2020self}
Trapit Bansal, Rishikesh Jha, Tsendsuren Munkhdalai, and Andrew McCallum. 2020.
\newblock Self-supervised meta-learning for few-shot natural language
  classification tasks.
\newblock In \emph{Proceedings of the 2020 Conference on Empirical Methods in
  Natural Language Processing (EMNLP)}, pages 522--534.

\bibitem[{Bender(2019)}]{bender2019benderrule}
Emily~M Bender. 2019.
\newblock The\# benderrule: On naming the languages we study and why it
  matters.
\newblock \emph{The Gradient}, 14.

\bibitem[{Bengio et~al.(1990)Bengio, Bengio, and Cloutier}]{bengio1990learning}
Yoshua Bengio, Samy Bengio, and Jocelyn Cloutier. 1990.
\newblock \emph{Learning a synaptic learning rule}.
\newblock Citeseer.

\bibitem[{Blasi et~al.(2021)Blasi, Anastasopoulos, and
  Neubig}]{blasi2021systematic}
Dami{\'a}n Blasi, Antonios Anastasopoulos, and Graham Neubig. 2021.
\newblock Systematic inequalities in language technology performance across the
  world's languages.
\newblock \emph{arXiv preprint arXiv:2110.06733}.

\bibitem[{Chi et~al.(2020{\natexlab{a}})Chi, Dong, Wei, Wang, Mao, and
  Huang}]{chi2020cross}
Zewen Chi, Li~Dong, Furu Wei, Wenhui Wang, Xian-Ling Mao, and Heyan Huang.
  2020{\natexlab{a}}.
\newblock Cross-lingual natural language generation via pre-training.
\newblock In \emph{Proceedings of the AAAI Conference on Artificial
  Intelligence}, volume~34, pages 7570--7577.

\bibitem[{Chi et~al.(2020{\natexlab{b}})Chi, Dong, Wei, Wang, Mao, and
  Huang}]{Chi_Dong_Wei_Wang_Mao_Huang_2020}
Zewen Chi, Li~Dong, Furu Wei, Wenhui Wang, Xian-Ling Mao, and Heyan Huang.
  2020{\natexlab{b}}.
\newblock Cross-lingual natural language generation via pre-training.
\newblock \emph{Proceedings of the AAAI Conference on Artificial Intelligence},
  34(05):7570--7577.

\bibitem[{Clark et~al.(2020)Clark, Choi, Collins, Garrette, Kwiatkowski,
  Nikolaev, and Palomaki}]{clark-etal-2020-tydi}
Jonathan~H. Clark, Eunsol Choi, Michael Collins, Dan Garrette, Tom Kwiatkowski,
  Vitaly Nikolaev, and Jennimaria Palomaki. 2020.
\newblock {T}y{D}i {QA}: A benchmark for information-seeking question answering
  in typologically diverse languages.
\newblock \emph{Transactions of the Association for Computational Linguistics},
  8:454--470.

\bibitem[{Dryer and Haspelmath(2013)}]{dryer2013world}
Matthew~S Dryer and Martin Haspelmath. 2013.
\newblock The world atlas of language structures online.

\bibitem[{Duan et~al.(2019)Duan, Yin, Zhang, Chen, and
  Luo}]{duan-etal-2019-zero}
Xiangyu Duan, Mingming Yin, Min Zhang, Boxing Chen, and Weihua Luo. 2019.
\newblock Zero-shot cross-lingual abstractive sentence summarization through
  teaching generation and attention.
\newblock In \emph{Proceedings of the 57th Annual Meeting of the Association
  for Computational Linguistics}, pages 3162--3172, Florence, Italy.
  Association for Computational Linguistics.

\bibitem[{Finn et~al.(2017)Finn, Abbeel, and Levine}]{finn2017model}
Chelsea Finn, Pieter Abbeel, and Sergey Levine. 2017.
\newblock Model-agnostic meta-learning for fast adaptation of deep networks.
\newblock In \emph{International Conference on Machine Learning}, pages
  1126--1135. PMLR.

\bibitem[{Gao et~al.(2019)Gao, Han, Zhu, Liu, Li, Sun, and
  Zhou}]{gao-etal-2019-fewrel}
Tianyu Gao, Xu~Han, Hao Zhu, Zhiyuan Liu, Peng Li, Maosong Sun, and Jie Zhou.
  2019.
\newblock {F}ew{R}el 2.0: Towards more challenging few-shot relation
  classification.
\newblock In \emph{Proceedings of the 2019 Conference on Empirical Methods in
  Natural Language Processing and the 9th International Joint Conference on
  Natural Language Processing (EMNLP-IJCNLP)}, pages 6251--6256, Hong Kong,
  China. Association for Computational Linguistics.

\bibitem[{Gu et~al.(2018)Gu, Wang, Chen, Li, and Cho}]{gu-etal-2018-meta}
Jiatao Gu, Yong Wang, Yun Chen, Victor O.~K. Li, and Kyunghyun Cho. 2018.
\newblock Meta-learning for low-resource neural machine translation.
\newblock In \emph{Proceedings of the 2018 Conference on Empirical Methods in
  Natural Language Processing}, pages 3622--3631, Brussels, Belgium.
  Association for Computational Linguistics.

\bibitem[{Hammarstr{\"o}m et~al.(2017)Hammarstr{\"o}m, Forkel, and
  Haspelmath}]{hammarstrom2017glottolog}
Harald Hammarstr{\"o}m, Robert Forkel, and Martin Haspelmath. 2017.
\newblock Glottolog 3.0 (max planck institute for the science of human history,
  jena).

\bibitem[{Hasan et~al.(2021)Hasan, Bhattacharjee, Islam, Mubasshir, Li, Kang,
  Rahman, and Shahriyar}]{hasan-etal-2021-xl}
Tahmid Hasan, Abhik Bhattacharjee, Md.~Saiful Islam, Kazi Mubasshir, Yuan-Fang
  Li, Yong-Bin Kang, M.~Sohel Rahman, and Rifat Shahriyar. 2021.
\newblock {XL}-sum: Large-scale multilingual abstractive summarization for 44
  languages.
\newblock In \emph{Findings of the Association for Computational Linguistics:
  ACL-IJCNLP 2021}, pages 4693--4703, Online. Association for Computational
  Linguistics.

\bibitem[{Houlsby et~al.(2019)Houlsby, Giurgiu, Jastrzebski, Morrone,
  De~Laroussilhe, Gesmundo, Attariyan, and Gelly}]{houlsby2019parameter}
Neil Houlsby, Andrei Giurgiu, Stanislaw Jastrzebski, Bruna Morrone, Quentin
  De~Laroussilhe, Andrea Gesmundo, Mona Attariyan, and Sylvain Gelly. 2019.
\newblock Parameter-efficient transfer learning for nlp.
\newblock In \emph{International Conference on Machine Learning}, pages
  2790--2799. PMLR.

\bibitem[{Hu et~al.(2020)Hu, Ruder, Siddhant, Neubig, Firat, and
  Johnson}]{hu2020xtreme}
Junjie Hu, Sebastian Ruder, Aditya Siddhant, Graham Neubig, Orhan Firat, and
  Melvin Johnson. 2020.
\newblock \href {http://arxiv.org/abs/2003.11080} {Xtreme: A massively
  multilingual multi-task benchmark for evaluating cross-lingual
  generalization}.
\newblock \emph{CoRR}, abs/2003.11080.

\bibitem[{Joshi et~al.(2020)Joshi, Santy, Budhiraja, Bali, and
  Choudhury}]{joshi-etal-2020-state}
Pratik Joshi, Sebastin Santy, Amar Budhiraja, Kalika Bali, and Monojit
  Choudhury. 2020.
\newblock The state and fate of linguistic diversity and inclusion in the {NLP}
  world.
\newblock In \emph{Proceedings of the 58th Annual Meeting of the Association
  for Computational Linguistics}, pages 6282--6293, Online. Association for
  Computational Linguistics.

\bibitem[{Koch et~al.(2015)Koch, Zemel, Salakhutdinov et~al.}]{koch2015siamese}
Gregory Koch, Richard Zemel, Ruslan Salakhutdinov, et~al. 2015.
\newblock Siamese neural networks for one-shot image recognition.
\newblock In \emph{ICML deep learning workshop}, volume~2. Lille.

\bibitem[{Kumar et~al.(2019)Kumar, Joshi, Mukherjee, Ramakrishnan, and
  Jyothi}]{kumar-etal-2019-cross}
Vishwajeet Kumar, Nitish Joshi, Arijit Mukherjee, Ganesh Ramakrishnan, and
  Preethi Jyothi. 2019.
\newblock Cross-lingual training for automatic question generation.
\newblock In \emph{Proceedings of the 57th Annual Meeting of the Association
  for Computational Linguistics}, pages 4863--4872, Florence, Italy.
  Association for Computational Linguistics.

\bibitem[{Ladhak et~al.(2020)Ladhak, Durmus, Cardie, and
  McKeown}]{ladhak-etal-2020-wikilingua}
Faisal Ladhak, Esin Durmus, Claire Cardie, and Kathleen McKeown. 2020.
\newblock {W}iki{L}ingua: A new benchmark dataset for cross-lingual abstractive
  summarization.
\newblock In \emph{Findings of the Association for Computational Linguistics:
  EMNLP 2020}, pages 4034--4048, Online. Association for Computational
  Linguistics.

\bibitem[{Lai et~al.(2019)Lai, Oguz, Yang, and
  Stoyanov}]{DBLP:journals/corr/abs-1909-07009}
Guokun Lai, Barlas Oguz, Yiming Yang, and Veselin Stoyanov. 2019.
\newblock \href {http://arxiv.org/abs/1909.07009} {Bridging the domain gap in
  cross-lingual document classification}.
\newblock \emph{CoRR}, abs/1909.07009.

\bibitem[{Lewis et~al.(2020{\natexlab{a}})Lewis, Ghazvininejad, Ghosh,
  Aghajanyan, Wang, and Zettlemoyer}]{lewis2020pretraining}
Mike Lewis, Marjan Ghazvininejad, Gargi Ghosh, Armen Aghajanyan, Sida Wang, and
  Luke Zettlemoyer. 2020{\natexlab{a}}.
\newblock \href {http://arxiv.org/abs/2006.15020} {Pre-training via
  paraphrasing}.

\bibitem[{Lewis et~al.(2020{\natexlab{b}})Lewis, Oguz, Rinott, Riedel, and
  Schwenk}]{lewis-etal-2020-mlqa}
Patrick Lewis, Barlas Oguz, Ruty Rinott, Sebastian Riedel, and Holger Schwenk.
  2020{\natexlab{b}}.
\newblock {MLQA}: Evaluating cross-lingual extractive question answering.
\newblock In \emph{Proceedings of the 58th Annual Meeting of the Association
  for Computational Linguistics}, pages 7315--7330, Online. Association for
  Computational Linguistics.

\bibitem[{Lin(2004)}]{lin-2004-rouge}
Chin-Yew Lin. 2004.
\newblock \href {https://aclanthology.org/W04-1013} {{ROUGE}: A package for
  automatic evaluation of summaries}.
\newblock In \emph{Text Summarization Branches Out}, pages 74--81, Barcelona,
  Spain. Association for Computational Linguistics.

\bibitem[{Lin et~al.(2019)Lin, Chen, Lee, Li, Zhang, Xia, Rijhwani, He, Zhang,
  Ma, Anastasopoulos, Littell, and Neubig}]{lin-etal-2019-choosing}
Yu-Hsiang Lin, Chian-Yu Chen, Jean Lee, Zirui Li, Yuyan Zhang, Mengzhou Xia,
  Shruti Rijhwani, Junxian He, Zhisong Zhang, Xuezhe Ma, Antonios
  Anastasopoulos, Patrick Littell, and Graham Neubig. 2019.
\newblock Choosing transfer languages for cross-lingual learning.
\newblock In \emph{Proceedings of the 57th Annual Meeting of the Association
  for Computational Linguistics}, pages 3125--3135, Florence, Italy.
  Association for Computational Linguistics.

\bibitem[{Littell et~al.(2017)Littell, Mortensen, Lin, Kairis, Turner, and
  Levin}]{littell-etal-2017-uriel}
Patrick Littell, David~R. Mortensen, Ke~Lin, Katherine Kairis, Carlisle Turner,
  and Lori Levin. 2017.
\newblock {URIEL} and lang2vec: Representing languages as typological,
  geographical, and phylogenetic vectors.
\newblock In \emph{Proceedings of the 15th Conference of the {E}uropean Chapter
  of the Association for Computational Linguistics: Volume 2, Short Papers},
  pages 8--14, Valencia, Spain. Association for Computational Linguistics.

\bibitem[{Malaviya et~al.(2017)Malaviya, Neubig, and
  Littell}]{malaviya-etal-2017-learning}
Chaitanya Malaviya, Graham Neubig, and Patrick Littell. 2017.
\newblock \href {https://doi.org/10.18653/v1/D17-1268} {Learning language
  representations for typology prediction}.
\newblock In \emph{Proceedings of the 2017 Conference on Empirical Methods in
  Natural Language Processing}, pages 2529--2535, Copenhagen, Denmark.
  Association for Computational Linguistics.

\bibitem[{Maurya et~al.(2021)Maurya, Desarkar, Kano, and
  Deepshikha}]{DBLP:conf/acl/MauryaDKD21}
Kaushal~Kumar Maurya, Maunendra~Sankar Desarkar, Yoshinobu Kano, and Kumari
  Deepshikha. 2021.
\newblock Zmbart: An unsupervised cross-lingual transfer framework for language
  generation.
\newblock In \emph{Findings of the Association for Computational Linguistics:
  {ACL/IJCNLP} 2021, Online Event, August 1-6, 2021}, volume {ACL/IJCNLP} 2021
  of \emph{Findings of {ACL}}, pages 2804--2818. Association for Computational
  Linguistics.

\bibitem[{M{'}hamdi et~al.(2021)M{'}hamdi, Kim, Dernoncourt, Bui, Ren, and
  May}]{mhamdi-etal-2021-x}
Meryem M{'}hamdi, Doo~Soon Kim, Franck Dernoncourt, Trung Bui, Xiang Ren, and
  Jonathan May. 2021.
\newblock {X}-{METRA}-{ADA}: Cross-lingual meta-transfer learning adaptation to
  natural language understanding and question answering.
\newblock In \emph{Proceedings of the 2021 Conference of the North American
  Chapter of the Association for Computational Linguistics: Human Language
  Technologies}, pages 3617--3632, Online. Association for Computational
  Linguistics.

\bibitem[{Mitra et~al.(2021)Mitra, Jain, Veerubhotla, and
  Gupta}]{mitra2021zero}
Rajarshee Mitra, Rhea Jain, Aditya~Srikanth Veerubhotla, and Manish Gupta.
  2021.
\newblock Zero-shot multi-lingual interrogative question generation for" people
  also ask" at bing.
\newblock In \emph{Proceedings of the 27th ACM SIGKDD Conference on Knowledge
  Discovery \& Data Mining}, pages 3414--3422.

\bibitem[{Murty et~al.(2021)Murty, Hashimoto, and Manning}]{murty2021dreca}
Shikhar Murty, Tatsunori Hashimoto, and Christopher~D Manning. 2021.
\newblock Dreca: A general task augmentation strategy for few-shot natural
  language inference.
\newblock In \emph{Proceedings of the 2021 Conference of the North American
  Chapter of the Association for Computational Linguistics: Human Language
  Technologies}, pages 1113--1125.

\bibitem[{Nie et~al.(2019)Nie, ge~Yao, Wang, Pan, and Lin}]{Nie2019ASR}
Feng Nie, Jin ge~Yao, Jinpeng Wang, Rong Pan, and Chin-Yew Lin. 2019.
\newblock A simple recipe towards reducing hallucination in neural surface
  realisation.
\newblock In \emph{ACL}.

\bibitem[{Nooralahzadeh et~al.(2020)Nooralahzadeh, Bekoulis, Bjerva, and
  Augenstein}]{DBLP:conf/emnlp/NooralahzadehBB20}
Farhad Nooralahzadeh, Giannis Bekoulis, Johannes Bjerva, and Isabelle
  Augenstein. 2020.
\newblock Zero-shot cross-lingual transfer with meta learning.
\newblock In \emph{Proceedings of the 2020 Conference on Empirical Methods in
  Natural Language Processing, {EMNLP} 2020, Online, November 16-20, 2020},
  pages 4547--4562. Association for Computational Linguistics.

\bibitem[{Oncevay et~al.(2020)Oncevay, Haddow, and
  Birch}]{oncevay-etal-2020-bridging}
Arturo Oncevay, Barry Haddow, and Alexandra Birch. 2020.
\newblock \href {https://doi.org/10.18653/v1/2020.emnlp-main.187} {Bridging
  linguistic typology and multilingual machine translation with multi-view
  language representations}.
\newblock In \emph{Proceedings of the 2020 Conference on Empirical Methods in
  Natural Language Processing (EMNLP)}, pages 2391--2406, Online. Association
  for Computational Linguistics.

\bibitem[{Papineni et~al.(2002)Papineni, Roukos, Ward, and
  Zhu}]{10.3115/1073083.1073135}
Kishore Papineni, Salim Roukos, Todd Ward, and Wei-Jing Zhu. 2002.
\newblock Bleu: A method for automatic evaluation of machine translation.
\newblock ACL '02, page 311–318, USA. Association for Computational
  Linguistics.

\bibitem[{Raffel et~al.(2020)Raffel, Shazeer, Roberts, Lee, Narang, Matena,
  Zhou, Li, and Liu}]{2020t5}
Colin Raffel, Noam Shazeer, Adam Roberts, Katherine Lee, Sharan Narang, Michael
  Matena, Yanqi Zhou, Wei Li, and Peter~J. Liu. 2020.
\newblock \href {http://jmlr.org/papers/v21/20-074.html} {Exploring the limits
  of transfer learning with a unified text-to-text transformer}.
\newblock \emph{Journal of Machine Learning Research}, 21(140):1--67.

\bibitem[{Rajpurkar et~al.(2016)Rajpurkar, Zhang, Lopyrev, and
  Liang}]{rajpurkar-etal-2016-squad}
Pranav Rajpurkar, Jian Zhang, Konstantin Lopyrev, and Percy Liang. 2016.
\newblock {SQ}u{AD}: 100,000+ questions for machine comprehension of text.
\newblock In \emph{Proceedings of the 2016 Conference on Empirical Methods in
  Natural Language Processing}, pages 2383--2392, Austin, Texas. Association
  for Computational Linguistics.

\bibitem[{Takahashi et~al.(2019)Takahashi, Shibata, Kawahara, and
  Kurohashi}]{takahashi-etal-2019-machine}
Norio Takahashi, Tomohide Shibata, Daisuke Kawahara, and Sadao Kurohashi. 2019.
\newblock Machine comprehension improves domain-specific {J}apanese
  predicate-argument structure analysis.
\newblock In \emph{Proceedings of the 2nd Workshop on Machine Reading for
  Question Answering}, pages 98--104, Hong Kong, China. Association for
  Computational Linguistics.

\bibitem[{Tarunesh et~al.(2021)Tarunesh, Khyalia, Kumar, Ramakrishnan, and
  Jyothi}]{DBLP:conf/eacl/TaruneshKKRJ21}
Ishan Tarunesh, Sushil Khyalia, Vishwajeet Kumar, Ganesh Ramakrishnan, and
  Preethi Jyothi. 2021.
\newblock Meta-learning for effective multi-task and multilingual modelling.
\newblock In \emph{Proceedings of the 16th Conference of the European Chapter
  of the Association for Computational Linguistics: Main Volume, {EACL} 2021,
  Online, April 19 - 23, 2021}, pages 3600--3612. Association for Computational
  Linguistics.

\bibitem[{van~der Heijden et~al.(2021)van~der Heijden, Yannakoudakis, Mishra,
  and Shutova}]{van-der-heijden-etal-2021-multilingual}
Niels van~der Heijden, Helen Yannakoudakis, Pushkar Mishra, and Ekaterina
  Shutova. 2021.
\newblock Multilingual and cross-lingual document classification: A
  meta-learning approach.
\newblock In \emph{Proceedings of the 16th Conference of the European Chapter
  of the Association for Computational Linguistics: Main Volume}, pages
  1966--1976, Online. Association for Computational Linguistics.

\bibitem[{Wan et~al.(2010)Wan, Li, and Xiao}]{wan2010cross}
Xiaojun Wan, Huiying Li, and Jianguo Xiao. 2010.
\newblock Cross-language document summarization based on machine translation
  quality prediction.
\newblock In \emph{Proceedings of the 48th Annual Meeting of the Association
  for Computational Linguistics}, pages 917--926.

\bibitem[{Wu et~al.(2020)Wu, Lin, Wang, Chen, Karlsson, Huang, and
  Lin}]{wu2020enhanced}
Qianhui Wu, Zijia Lin, Guoxin Wang, Hui Chen, B{\"o}rje~F Karlsson, Biqing
  Huang, and Chin-Yew Lin. 2020.
\newblock Enhanced meta-learning for cross-lingual named entity recognition
  with minimal resources.
\newblock In \emph{Proceedings of the AAAI Conference on Artificial
  Intelligence}, volume~34, pages 9274--9281.

\bibitem[{Xue et~al.(2021)Xue, Constant, Roberts, Kale, Al-Rfou, Siddhant,
  Barua, and Raffel}]{xue-etal-2021-mt5}
Linting Xue, Noah Constant, Adam Roberts, Mihir Kale, Rami Al-Rfou, Aditya
  Siddhant, Aditya Barua, and Colin Raffel. 2021.
\newblock \href {https://doi.org/10.18653/v1/2021.naacl-main.41} {m{T}5: A
  massively multilingual pre-trained text-to-text transformer}.
\newblock In \emph{Proceedings of the 2021 Conference of the North American
  Chapter of the Association for Computational Linguistics: Human Language
  Technologies}, pages 483--498, Online. Association for Computational
  Linguistics.

\bibitem[{Yan et~al.(2020)Yan, Zhang, Jin, and Zhou}]{yan2020multi}
Ming Yan, Hao Zhang, Di~Jin, and Joey~Tianyi Zhou. 2020.
\newblock Multi-source meta transfer for low resource multiple-choice question
  answering.
\newblock In \emph{Proceedings of the 58th Annual Meeting of the Association
  for Computational Linguistics}, pages 7331--7341.

\end{thebibliography}
\bibliographystyle{acl_natbib}


\newpage
\newpage
\appendix

\section*{Appendices}


\section{Evaluation Metric Setting}
\label{sec:app_humEval}
We use the multilingual version of ROUGE released by \citet{hasan-etal-2021-xl} where they use language-specific tokenizers and stemmers. Inspired by this, we also added the language-specific tokenizer in sacreBLEU implementation to compute BLEU Score. 

\section{Miscellaneous}
\begin{enumerate}
 \item To the best of our knowledge, this is the first study towards meta-learning for the cross-lingual generation. The recent publications of applied meta-learning in NLP are listed here: \url{https://jeffeuxmartin.github.io/meta-learning-hlp/} and \url{https://github.com/ha-lins/MetaLearning4NLP-Papers}  (accessed on 15th March, 2022)
    \item In the proposed framework the data instance tag is <fxx><2xx>, where <fxx> is tag for input document language and \textit{<2xx>} for target language for example: <fen> <2en>. In this work, the input and target document languages are the same. The tag will be easily adapted in the future, where input and target document languages are different. For example, the tag \textit{<fen> <2fr>} indicates that the input document language is English (en) and target document language is French (fr). 
    \item We are aware that, recently adapter modules \cite{houlsby2019parameter} have emerged as alternate solution for catastrophic forgetting problem. In future, we will compare Meta-X\textsubscript{NLG} performance with   Meta-X\textsubscript{NLG} + adapters model.  
    \item The additional Tamil language in TyDiQA is taken from Kaggle\footnote{  \url{https://www.kaggle.com/c/chaii-hindi-and-tamil-question-answering/data}}
\end{enumerate}
\section{Other Details}
\label{app:lanc_cd}
\begin{table*}[!htb]
\centering
\scalebox{0.7}{
\begin{tabular}{|c|l|c|c|c|c|c|c|c|c|}
    \hline\hline
 \textbf{SN} & \textbf{Language} &  \textbf{ISO-2}  &  \textbf{ISO-3}  &   \textbf{Adap. PT}  &   \textbf{XL-Sum}  &  \textbf{Wikilingua } & \textbf{MLQA***} & \textbf{TyDiQA****} & \textbf{XQuAD***}      \\
 & & & & \textbf{train/valid/test} & \textbf{test} & \textbf{test} & \textbf{test} & \textbf{test} &\textbf{ test}   \\
\hline \hline
1 & English* & en & eng & 5k/1k/1k & 300k/11k/11k & 100k/13k/28k& 90k/10k/11k & 90k/10k/11k & 90k/10k/11k\\
\hline
2 & Hindi & hi & hin & 5k/1k/1k & 8847 & 1983 & 4918 & -& 1190\\
\hline
3 & Urdu & ur & urd & 5k/1k/1k & 8458 &- & -&- & -\\
\hline
4 & Telugu & te & tel & 5k/1k/1k & 1302 & 899 & -& 5563 & -\\
\hline
5 & Turkish & tr & tru & 5k/1k/1k & 3397 & -&- & -& 1190\\
\hline
6 & Finnish & fi & fin & 5k/1k/1k & - &- & -& 6855 & -\\
\hline
7 & Japanese & ja & jpn & 5k/1k/1k &  889 & 2529 & 5000** & -& -\\
\hline
8 & Korean & ko & kor & 5k/1k/1k & 550 & 2435 & -& 1620& -\\
\hline
9 & Gujarati & gu & guj & 5k/1k/1k & 1139 &- &- & -& -\\
\hline
10 & Bengali & bn & ben & 5k/1k/1k & 1012 &- &- & 2390 & -\\
\hline
11 & Marathi & mr & mar & 5k/1k/1k & 1362 &- & -& -& -\\
\hline
12 & Nepali & np & nep & 5k/1k/1k & 725 &- &- & -& -\\
\hline
13 & Tamil & ta & tam & 5k/1k/1k & 2027 & -&- & 368** & -\\
\hline
14 & Punjabi & pa & pan & 5k/1k/1k & 1026 &- &- &- & -\\
\hline
15 & Swahili & sw & swa & 5k/1k/1k & 987 &- &- & 2755 & -\\
\hline
16 & Spanish & es & spa & 5k/1k/1k & 4763 & 22626 & 5253 & - & 1190\\
\hline
17 &  Italian & it & ita & 5k/1k/1k & - & 10187 &- & -& -\\
\hline
18 & Portuguese & pt & por & 5k/1k/1k & 7175 & 16326 &- & -& -\\
\hline
19 & Romanian & ro & ron & 5k/1k/1k & - & -&- & -& 1190 -\\
\hline
20 & Dutch & nl & nld & 5k/1k/1k & - & 6248 &- & -& -\\
\hline
21 & German & de & deu & 5k/1k/1k & - & 11667 & 4517 & -& 1190\\
\hline
22 & French & fr & fra & 5k/1k/1k & 1086 & 12728 & -& -& -\\
\hline
23 & Russian & ru & rus & 5k/1k/1k & 7780& 10577 &- & 6490 & 1190 \\
\hline
24 & Czech & cs & ces & 5k/1k/1k & - & 1438 &- & -& -\\
\hline
25 & Vietnamese & vi & vie & 5k/1k/1k & 4013 & 3916 & 5459 & -& 1190\\
\hline
26 & Thai  & th & tha & 5k/1k/1k & 826& 2949 & -&-& 1190 \\
\hline
27 & Chinese (Sim) & zh & zho & 5k/1k/1k & 4670 & 3772 & 5137 & -& 1190\\
\hline
28 & Indonesian & id & ind & 5k/1k/1k  & 4780 & 9495 & -& 5702 & -\\
\hline
29 & Greek & el & ell & 5k/1k/1k & - & - & - & -& 1190\\
\hline
30 & Arabic & ar & ara & 5k/1k/1k & 4689 & 5840 & 5335 & 14805 & 1190\\
\hline
\end{tabular}
}
\caption{Details of the datasets used in Meta-X\textsubscript{NLG}. For adaptive pre-training small 5k/1k/1k dataset is used. *-English is a high resource language for which all three splits were used, as shown in Row 1. **-additional language added in the dataset. ***-dataset does not have validation split, so a test data set of centroid languages is used in training.****-TyDiQA does not have a test set, so the training set is used for evaluation (test set).}
\label{tab:data_stats}
\end{table*}

\begin{table*}[!htb]
\centering
\scalebox{0.8}{
\begin{tabular}{l|lc|lc|lc}
    \hline\hline
    \textbf{Dataset}    & \multicolumn{2}{c|}{\textbf{1st Centroid Lang}}
            & \multicolumn{2}{c|}{\textbf{2nd Centroid Lang}}
                    & \multicolumn{2}{c}{\textbf{3rd Centroid Lang}}        \\
   &   \textbf{Lang } &  \textbf{Val Size}  &  \textbf{ Lang}  &   \textbf{Val Size}  &   \textbf{Lang}  &    \textbf{Val Size}      \\
    \hline 
 XL-Sum  & Punjabi & 1026  &Spanish & 1026 & Vietnamese & 1026\\
 Wikilingua  & Korean & 1011 & Spanish & 1011 & Vietnamese & 1011 \\
 MLQA  & Japanese & 4517 & German & 4517 & Vietnamese & 4517 \\
 TyDiQA  & Telugu & 5562 & - & - & Arabic & 5562\\
 XQuAD  & Turkish & 1190 & Spanish & 1190 & Thai & 1190\\
\hline \hline
\end{tabular} 
}
\caption{\small Size of centroid languages validation set used in the proposed Meta-X\textsubscript{NLG} framework. The same number of examples are sampled from each centroid language.}

\label{tab:mtrain_valdsize}
\end{table*}

\begin{figure*}[!htb]
    \centering
    \includegraphics[width=15cm]{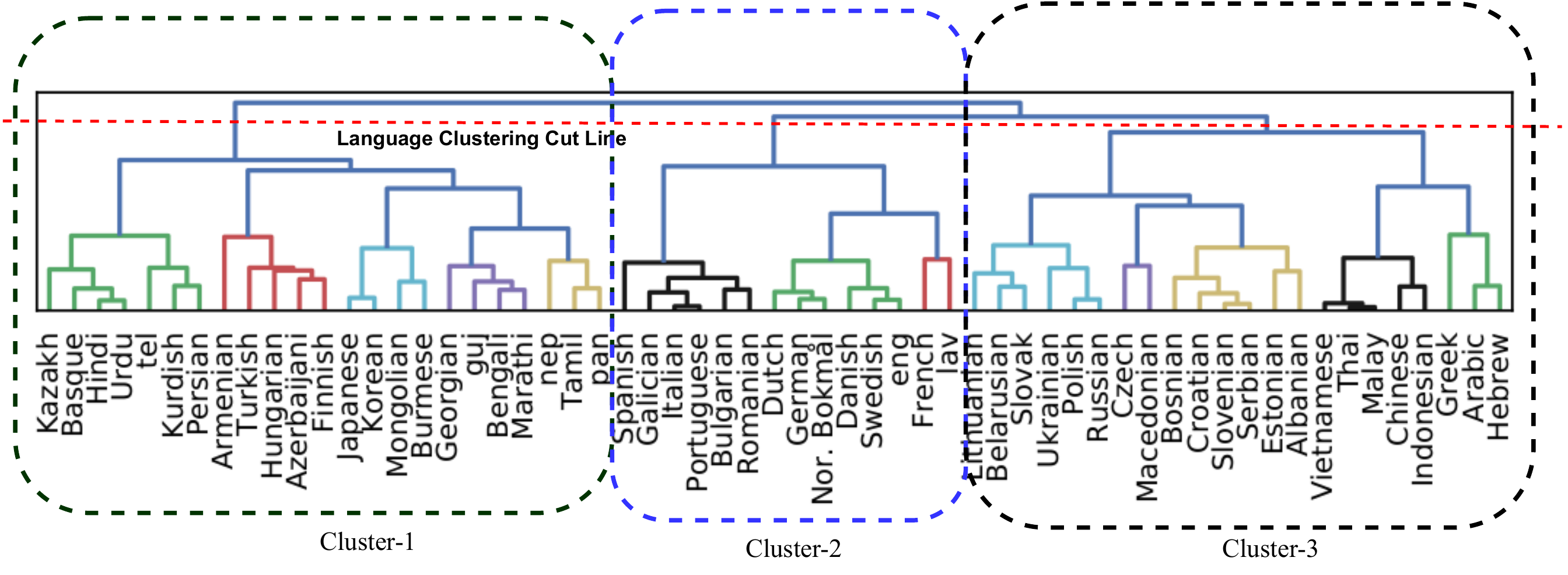}
    \caption{Language clustering based on multi-view representation proposed by \citet{oncevay-etal-2020-bridging}. We intentionally show more than 30 languages in the clustering, which will be useful for scaling the proposed work in the future. As per our application need,  we added multiple languages in clustering over originally proposed by the authors. Additional languages are: Telugu (tel), Gujarati (guj), Nepali (nep), Punjabi (pan), English (eng).}
    \label{fig:meta_xnlg_lang_cluster}
\end{figure*}

\begin{table*}[!htb]
\centering
\scalebox{0.7}{
\begin{tabular}{|l|lc|lc|lc|l|l|}
    \hline\hline
\textbf{Task/Dataset}    & \multicolumn{2}{c|}{\textbf{Cluster-1}}
            & \multicolumn{2}{c|}{\textbf{Cluster-2}}
                    & \multicolumn{2}{c|}{\textbf{Cluster-3}}
                            &    \textbf{Centroid Lang}  &    \textbf{Non-Centroid Lang}             \\
    \hline
   &   \textbf{Lang } &  \textbf{MeanCD}  &  \textbf{ Lang}  &   \textbf{MeanCD}  &   \textbf{Lang}  &    \textbf{MeanCD }&   \textbf{Meta-train Lang} &   \textbf{Target Lang}   \\
    \hline \hline
Sum/XL-Sum  &   Punjabi  &  0.505    &  Spanish  &   0.253  &   Vietnamese &   0.291 &   Punjabi & Tamil ,Marathi     \\
            &   Tamil    &   0.547    &   Portuguese    &  0.437     &  Thai     &   0.326    &   Spanish & Gujarati , Bengali  \\
            &   Marathi    &   0.548    &   French     &   0.477    &  Indonesian     &  0.327     & Vietnamese  & Telugu, Hindi \\
            &   Gujarati    &  0.550    &       &       &   Arabic     &  0.465     &   &  Nepali , Urdu         \\
            &   Bengali   &   0.566    &       &       &   Chinese    &   0.561     &   &   Japanese, Turkish        \\
            &   Telugu   &   0.574     &       &       &  Russian     &   0.902    &    &   Korean, Swahili       \\
            &   Hindi    &  0.630    &       &       &       &       &      &  Portuguese, French       \\
            &   Nepali    &   0.659    &       &       &       &       &    &  Thai, Indonesian          \\
            &   Urdu       &  0.663     &       &       &       &       &   &  Arabic, Chinese   \\
            &   Japanese    &  0.749     &       &       &       &       &  & Russian           \\
            &   Turkish    &  0.803     &       &       &       &       &    &          \\
            &   Korean    &   0.808    &       &       &       &       &     &         \\
            &   Swahili   &   -    &       &       &       &       &         &     \\
    \hline
Sum/Wikilingua  &  Korean  &   0.558    &  Spanish     &   0.459 & Vietnamese  &  0.484  &   Korean &  Japanese, Turkish          \\
            &  Japanese   &   0.583    &  French     &    0.476   &    Thai   &  0.496     &      Spanish & Hindi, French        \\
            &  Turkish& 0.620  &  German   &   0.529    &  Indonesian     &  0.536     &   Vietnamese  &  German, Portuguese        \\
            &  Hindi   &   1.166   &  Portuguese     &   0.535    &  Arabic     &   0.595    &    &  Italian, Dutch  \\
            &      &      &          Italian    &   0.566    &   Chinese    &     0.758   &  &  Thai,  Indonesian  \\
            &     &       &          Dutch    &   0.674    &   Russian    &    0.897   &   &  Arabic, Chinese  \\
            &     &       &             &       &  Czech    &     1.374     &   & Russian, Czech  \\
    \hline
QG/MLQA  &  Japanese &    1.156   &   German    &   0.843    &  Vietnamese     &   0.299     &   Japanese  & Hindi,  Spanish       \\
         &  Hindi  &    1.156   &  Spanish  & 0.843  &   Chinese    &      0.459     &    German    &   Chinese, Arabic   \\
         &   &       &       &       &   Arabic    &  0.483      &   Vietnamese  &       \\
    \hline
QG/TyDiQA  & Telugu  &  0.682  &      &      &   Arabic   &   0.579     &     Telugu   &   Tamil, Bengali  \\
&  Tamil  &  0.719    &      &      &   Indonesian   &   0.619     &     Arabic &  Finnish, Korean     \\
&  Bengali &  0.769    &      &      &    Russian   &     0.940  &     & Swahili, Indonesian         \\
& Finnish  &  0.785    &      &      &      &        &  & Russian           \\
& Korean  &  0.828    &      &      &      &        &   &          \\
& Swahili &   -   &      &      &      &        &   &          \\
\hline
QG/XQuAD  & Turkish  & 1.038   &  Spanish    &   0.606   &  Thai    &    0.515    &  Turkish   & Hindi, Romanian         \\
& Hindi &  1.038  &   Romanian   &  0.788    &  Arabic    &    0.516    &    Spanish &  German, Arabic        \\
&  &    &   German   &    1.024   &   Vietnamese   &    0.519    &     Thai & Vietnamese,  Chinese      \\
&  &    &      &      &  Chinese    &    0.813    &       & Russian, Greek      \\
&  &    &      &      &   Russian   &   0.926     &     &        \\
&  &    &      &      &   Greek   &    1.071    &       &      \\
\hline \hline
\end{tabular}
}
\caption{ \small Details of language clustering for each dataset, mean cosine distance (meanCD), and centroid languages. For each dataset, we group languages into three clusters as shown in Figure \ref{fig:ov_meta_xnlg}. The Swahili language does not have any typological or task-based representations, so we added it to cluster 1 based on language typological features and heuristics. For the TyDiQA dataset, only two clusters are obtained as cluster-2 does not have any language. If a cluster has only two languages, we randomly selected any language as centroid language.}
\label{tab:lang_cluster}
\end{table*}

\begin{table*}[!htb]
\centering
\scalebox{0.55}{
\begin{tabular}{l|ccc|ccc|ccc}
    \hline\hline
\textbf{Setup} & & \textbf{English (Supervised)}& & & \textbf{Hindi  (Zero-shot)} & & & \textbf{Bengali (Zero-shot)} &\\
 & \textbf{R-1} & \textbf{R-2} & \textbf{R-L} &  \textbf{R-1} & \textbf{R-2} & \textbf{R-L} &  \textbf{R-1} & \textbf{R-2} & \textbf{R-L} \\
 \hline       
Without Adaptive Pre-training Step & 36.05 & 13.87 & 28.34 & 00.32 & 00.06 & 00.32 & 00.13 & 00.00 & 00.13 \\
Joint Training (T5 PTObj + EngFT [1:100]) \cite{xue-etal-2021-mt5} & 34.19 & 12.09 & 26.47 & 22.02 & 06.03 & 18.60 & 13.76 & 03.64 & 12.32\\
randSum Objective followed by EngFT \cite{DBLP:conf/acl/MauryaDKD21} & 33.38 & 11.57 & 26.00 & 24.31 & 07.11 & 19.91 & 16.23 & 04.32 & 14.66\\
T5 PTObj  followed by EngFT (proposed) & 34.15 & 11.99 & 26.59 & 26.75 & 08.39 & 22.24 & 18.63 & 05.71 & 16.12 \\
\hline \hline
\end{tabular} 
}
\caption{\small Results with different adaptive pre-training objectives. mT5 is a base pre-trained model for above all experimental setups. T5 PTObj is the T5 model's pre-training objective proposed by \citet{2020t5}. EngFT is English fine-tuning of base/adaptive pre-trained model. The results are shown on selected languages with XL-Sum dataset in standard supervised fine-tuning (English) and zero-shot setting (Hindi and Bengali). Proposed adaptive pre-training outperforms existing approaches for zero shot transfer.}

\label{tab:adpobj}
\end{table*}

\begin{table*}
\centering
\scalebox{0.5}{
\begin{tabular}{c|l|c|c|c|c|c|c|c|c|c|c|c|c|c|c|c|c|c|c|c|c}
    \hline\hline
\textbf{SetUp} & \textbf{Meta-Train Langs} & \textbf{fr} &  \textbf{gu}  &  \textbf{id}  &   \textbf{th}  &   \textbf{ta}  &  \textbf{hi} & \textbf{mr} & \textbf{ja} &  \textbf{ko}&  \textbf{tr}&  \textbf{ru}&  \textbf{sw}&  \textbf{pt}&  \textbf{ar}&  \textbf{te}&  \textbf{ur}&  \textbf{ne}&  \textbf{bn}&  \textbf{zh} &\textbf{avg}\\
 \hline
1* & pa & 16.59 & 7.55 & 15.87 & 23.57 & 11.10 & 13.22 & 9.54 & 24.17 & 17.67 & 15.61 & 13.51 & 17.34 & 16.42 & 15.94 & 9.19 & 12.69 & 11.84 & 13.25 & 20.71 & 15.04 \\
2* & es & 21.35 & 12.73 & 19.54 & 23.82 & 10.42 & 18.77 & 10.99 & 24.15 & 18.02 & 15.87 & 14.10 & 20.03 & 19.72 & 17.46 & 10.13 & 20.12 & 15.06 & 16.00 & 22.01 & 17.38 \\
3* & vi & 19.67 & 12.34 & 18.69 & 25.02 & 11.05 & 19.41 & 10.90 & 23.77 & 18.46 & 15.15 & 14.56 & 20.40 & 18.02 & 17.43 & 10.69 & 20.23 & 14.42 & 15.47 & 21.58 & 17.22 \\
4*  & ru & 17.60 & 12.89 & 16.97 & 23.54 & 10.50 & 18.03 & 10.75 & 24.28 & 18.09 & 16.36 & - & 18.25 & 17.32 & 17.63 & 10.44 & 20.52 & 14.28 & 14.40 & 22.18 & 16.89 \\
5*  & tr &  16.57 & 12.83 & 16.04 & 23.77 & 10.10 & 17.72 & 10.65 & 24.06 & 17.01 & - & 14.90 & 19.46 & 17.34 & 17.59 & 10.40 & 20.12 & 13.51 & 13.35 & 21.01 & 16.46 \\
6**  & np & 16.89 & 9.23 & 16.47 & 23.44 & 10.70 & 21.51 & 10.45 & 24.73 & 17.12 & 15.28 & 14.16 & 17.03 & 16.54 & 16.03 & 10.43 & 19.21 & - & 13.28 & 21.81 & 16.35 \\
7**  & th & 17.86 & 11.60 & 17.25 & - & 10.78 & 17.98 & 10.30 & 21.07 & 17.89 & 15.73 & 14.48 & 18.16 & 17.59 & 17.19 & 9.87 & 20.11 & 13.56 & 15.65 & 15.35 & 15.69 \\
8*  & vi, pa & 19.50 & 7.98 & 18.02 & 24.41 & 11.25 & 13.33 & 9.45 & 23.96 & 17.37 & 15.09 & 13.61 & 19.34 & 17.99 & 16.13 & 9.11 & 14.05 & 11.93 & 13.20 & 18.91 & 15.51 \\
8* & tr, es & 21.40 & 12.55 & 19.73 & 23.75 & 11.65 & 20.61 & 10.71 & 24.92 & 19.28 & - & 14.12 & 20.11 & 19.44 & 17.17 & 11.74 & 21.40 & 14.78 & 16.54 & 22.82 & 17.93 \\
10* & fr, vi &  - & 12.49 & 19.51 & 23.72 & 11.12 & 18.83 & 10.38 & 24.01 & 18.74 & 15.98 & 14.01 & 19.40 & 18.96 & 17.18 & 10.52 & 20.44 & 14.32 & 15.19 & 22.36 & 17.06 \\
11**  & ur, zh & 18.06 & 12.56 & 17.26 & 22.30 & 11.95 & 14.27 & 11.53 & 21.40 & 18.51 & 17.02 & 14.73 & 17.58 & 17.20 & 17.76 & 11.18 & - & 14.41 & 15.98 & - & 16.10 \\
12**  & th, pt & 21.28 & 12.39 & 19.60 & - & 10.83 & 17.90 & 10.04 & 22.49 & 17.02 & 16.07 & 14.52 & 20.19 & - & 17.61 & 10.00 & 19.79 & 13.77 & 15.10 & 21.45 & 16.47 \\
13@  & pa, pt &  21.13 & 8.72 & 19.92 & 23.89 & 11.64 & 14.38 & 9.65 & 24.13 & 17.36 & 16.89 & 14.91 & 20.90 & - & 17.36 & 9.95 & 15.53 & 11.66 & 13.37 & 22.04 & 16.30\\
14@  & es, bn & 21.61 & 10.53 & 18.85 & 23.23 & 11.06 & 17.33 & 10.15 & 24.31 & 17.25 & 15.68 & 13.69 & 19.32 & 19.27 & 16.29 & 10.46 & 20.40 & 11.75 & - & 19.48 & 16.70\\
15*  & pa,fr,ru & - & 9.80 & 19.17 & 23.39 & 10.54 & 13.97 & 9.43 & 24.41 & 17.50 & 16.56 & - & 19.52 & 19.07 & 16.08 & 9.03 & 16.44 & 11.43 & 13.01 & 21.71 & 15.95\\
16*  & pa,es,ru & 21.34 & 9.42 & 19.04 & 24.58 & 10.67 & 13.17 & 9.02 & 24.04 & 16.92 & 16.30 & - & 19.90 & 19.60 & 16.20 & 8.98 & 14.97 & 11.86 & 12.76 & 21.89 & 16.15 \\
17*  & vi,pa,fr & - & 9.75 & 19.31 & 23.65 & 11.18 & 13.98 & 9.41 & 24.52 & 17.91 & 15.88 & 13.79 & 20.20 & 19.24 & 16.28 & 9.47 & 15.68 & 11.78 & 13.75 & 19.48 & 15.85 \\
18**  & ko,pt,th & 21.66 & 12.94 & 19.93 & - & 11.94 & 20.35 & 10.42 & 24.46 & - & 17.99 & 15.55 & 21.22 & - & 18.58 & 11.23 & 21.54 & 15.20 & 16.06 & 16.72 & 17.24 \\
19**  & gu,pt,ar & 21.83 & - & 19.52 & 23.74 & 10.30 & 14.46 & 7.71 & 23.51 & 15.57 & 15.34 & 13.73 & 19.40 & - & - & 9.62 & 18.77 & 11.30 & 12.88 & 21.03 & 16.17 \\
20@  & es,th,ar & 22.11 & 12.14 & 19.60 & - & 10.60 & 17.22 & 9.92 & 22.88 & 16.78 & 16.18 & 13.81 & 20.42 & 20.09 & - & 10.25 & 19.55 & 13.58 & 15.35 & 17.27 & 16.34 \\
21@  & pa,pt,vi & 21.75 & 9.65 & 19.80 & 24.49 & 11.41 & 13.82 & 9.81 & 24.51 & 17.70 & 16.16 & 14.55 & 20.39 & - & 17.28 & 10.04 & 15.71 & 11.70 & 13.91 & 20.97 & 16.31 \\
22*  & pa,es,vi,fr & - & 9.35 & 19.74 & 23.91 & 11.11 & 13.86 & 8.96 & 24.82 & 17.70 & 16.54 & 13.57 & 20.65 & 20.16 & 16.43 & 9.52 & 16.76 & 11.73 & 13.48 & 19.81 & 16.01 \\
23*  & pa,ep,vi,ru & 21.90 & 8.39 & 19.28 & \textbf{24.89} & 10.65 & 14.19 & 9.38 & 24.25 & 16.47 & 16.00 & - & 21.20 & 20.12 & 16.38 & 9.19 & 16.07 & 11.62 & 12.98 & 19.03 & 16.06 \\
24*  & pa,es,vi, tr& 22.35 & 9.89 & 20.57 & 24.59 & 11.45 & 15.10 & 9.59 & 25.44 & 17.70 & - & 13.89 & 21.55 & 20.28 & 17.23 & 10.00 & 17.20 & 12.73 & 13.58 & 19.82 & 16.83 \\
25**  & zh,bn,te,pt & 21.73 & 10.94 & 18.98 & 22.99 & 10.58 & 16.23 & 9.46 & 20.57 & 16.16 & 15.80 & 13.57 & 20.23 & - & 16.23 & - & 19.51 & 12.23 & - & - & 16.35 \\
26**  & id,sw,ur,pt & 22.70 & 12.77 & - & 24.17 & 10.95 & 15.94 & 10.68 & 24.77 & 17.58 & 17.13 & 14.42 & - & - & 18.64 & 10.39 & - & 13.70 & 14.40 & 22.87 & 16.74 \\
27@  & pa,es,vi,hi & 21.81 & 8.66 & 19.21 & 24.43 & 10.64 & - & 11.03 & 24.25 & 17.20 & 16.12 & 12.89 & 20.86 & 19.93 & 16.25 & 9.58 & 16.15 & \textbf{16.36} & 12.56 & 13.78 & 16.21 \\
28@  & pa,es,vi,ko & 22.33 & 12.47 & 20.70 & 23.70 & 12.53 & 19.55 & 10.75 & 25.44 & - & 17.90 & 15.02 & \textbf{22.63} & 19.97 & 18.33 & 11.68 & 21.52 & 14.71 & 16.26 & 21.32 & 18.16 \\
29*  & pa,es,vi,fr,tr & - & 10.26 & 20.39 & 24.04 & 11.12 & 14.79 & 9.08 & 25.42 & 17.75 & - & 13.35 & 21.17 & 20.28 & 16.50 & 9.65 & 17.43 & 12.43 & 14.01 & 20.62 & 16.37 \\
30*  & pa,es,vi,ru,mr & 21.77 & 10.12 & 19.44 & 23.85 & 10.81 & 23.85 & - & 24.20 & 16.95 & 16.02 & - & 20.60 & 19.97 & 16.30 & 9.57 & 17.46 & 15.71 & 13.47 & 18.40 & 17.56 \\
31**  & id,sw,ur,po,te & 22.43 & 11.19 & - & 23.88 & 9.87 & 16.08 & 9.64 & 24.21 & 16.05 & 17.05 & 14.19 & - & - & 18.54 & - & - & 13.08 & 13.19 & 20.44 & 16.42 \\
32@  & pa,es,te,mr,gu & 20.51 & - & 18.05 & 22.01 & 9.69 & \textbf{23.94} & - & 21.93 & 15.32 & 15.04 & 11.83 & 18.51 & 19.39 & 14.60 & - & 16.70 & 15.81 & 12.70 & 10.13 & 16.63 \\
33*  & pa,es,vi,fr,tr,ru & - & 9.98 & 20.59 & 24.61 & 11.14 & 14.72 & 9.21 & 25.18 & 17.53 & - & - & 21.54 & 20.55 & 16.61 & 9.65 & 17.72 & 12.07 & 13.71 & 21.80 & 16.66 \\
34*  & pa,es,vi,fr,tr,ru,mr & - & 10.15 & 20.65 & 24.42 & 10.56 & 24.34 & - & 24.66 & 17.09 & - & - & 21.28 & 20.60 & 16.11 & 9.97 & 18.21 & 15.81 & 13.21 & 19.32 & 17.76 \\
35*  & pa,es,vi,fr,tr,ru,mr,ja & - & 9.88 & 19.61 & 23.51 & 9.83 & 23.40 & - & - & 13.27 & - & - & 21.43 & \textbf{20.36} & 15.83 & 9.24 & 15.66 & 16.24 & 12.68 & 20.32 & 16.52 \\
36* & \textbf{Meta-X\textsubscript{NLG}(pa,es,vi)}& \textbf{22.83} & \textbf{14.02} & \textbf{21.54} & 24.61 & \textbf{12.88} & 23.09 & \textbf{12.58} & 25.33 & \textbf{20.12} & \textbf{18.65} & \textbf{17.31} & \textbf{22.63}  & 20.24 & \textbf{20.11} & \textbf{12.07} & \textbf{23.41} & 15.45 & \textbf{17.96} & \textbf{22.95} & \textbf{19.40} \\
\hline \hline
\end{tabular} 
}
\caption{\small  Meta-X\textsubscript{NLG}'s zero-shot evaluation scores (Rouge-L) with different meta-training language combinations on the XL-Sum dataset. We cut the hierarchical clustering dendogram shown in Figure \ref{fig:meta_xnlg_lang_cluster}, at the lower level to obtain more clusters. In total, we obtain eight centroid languages, i.e., pa, es, vi, tr, ja, mr, fr and ru. '-' indicates the language used in training, so scores are not zero-shot and not included. Markers '*', '**', and '@' indicate meta training with all-centroid, all-non-centroid, and mix of both (centroid \& non-centroid) languages.}
\label{tab:xlsum_rslt_anays1}
\end{table*}

\begin{figure*}[!htb]
    \centering
    \includegraphics[width=15cm]{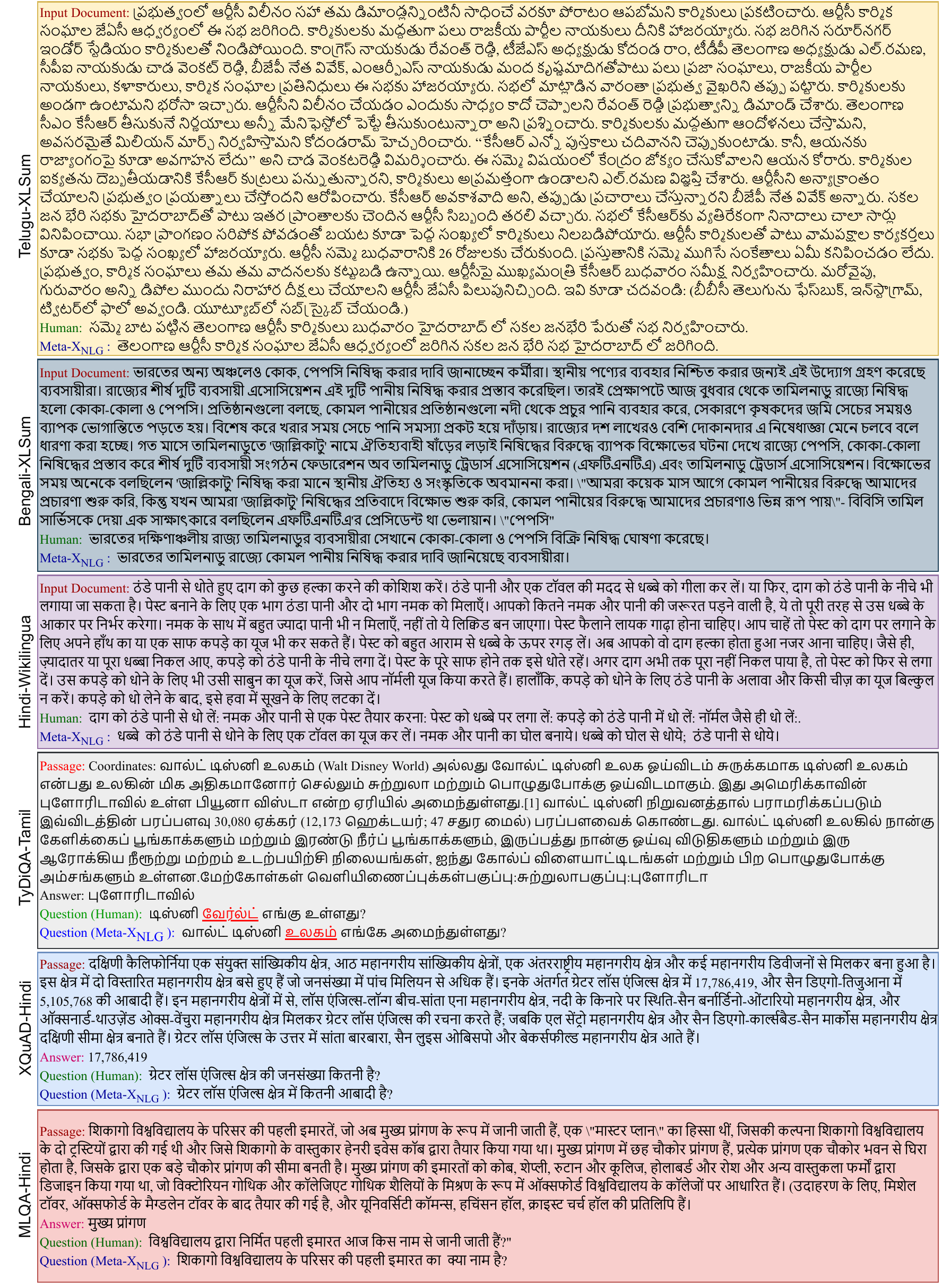}
    \caption{Zero-shot samples generated by Meta-X\textsubscript{NLG} in Telugu, Tamil, Bengali and Hindi languages. The top three samples are for ATS and the bottom three are for QG tasks. The generated samples are taken from all five datasets. In some instances, the model learns to generate an actual target language script even though the reference is in transliterated form. See the underlined token (in red font color) in the TyDiQA-Tamil example.}
    \label{fig:meta_xnlg_sample}
\end{figure*}

\end{document}